\documentclass[conference]{IEEEtran}
\IEEEoverridecommandlockouts
\usepackage{cite}
\usepackage{amsmath,amssymb,amsfonts}
\usepackage{algorithmic}
\usepackage{graphicx}
\usepackage{textcomp}
\usepackage{xcolor}
\usepackage{comment}
\usepackage{amsthm} 
\usepackage{makecell}

\newtheorem{definition}{Definition}

\newtheoremstyle{exampleStyle}
  {5pt}   
  {5pt}   
  {\itshape} 
  {}        
  {\bfseries} 
  {.}       
  { }       
  {}        

\theoremstyle{exampleStyle}
\newtheorem{example}{Example}

\usepackage{enumitem}
\usepackage{amsthm}
\usepackage{multirow}
\usepackage{booktabs}
\usepackage[caption=false]{subfig}
\usepackage{xspace}
\usepackage[table]{xcolor}
\usepackage[ruled,linesnumbered,noend]{algorithm2e}

\usepackage[hidelinks]{hyperref}

\def\BibTeX{{\rm B\kern-.05em{\sc i\kern-.025em b}\kern-.08em
    T\kern-.1667em\lower.7ex\hbox{E}\kern-.125emX}}
\begin{document}
\newcommand{\cond}{condensation}
\newcommand{\method}{STemDist\xspace}
\newcommand{\reduction}{DD reduction}
\newcommand{\smallsection}[1]{\noindent\underline{\smash{\textbf{#1:}}}}
\newcommand{\red}[1]{\textcolor{red}{#1}}
\newcommand{\blue}[1]{{\color{blue}#1}}
\newcommand{\green}[1]{{\color{green}#1}}
\newcommand{\setofinput}{\mathcal{X}_\mathcal{T}}
\newcommand{\setoffuture}{\mathcal{Y}_\mathcal{T}}

\newcommand{\setofinputmat}{\mathcal{X}^{mat}_\mathcal{T}}
\newcommand{\setoffuturemat}{\mathcal{Y}^{mat}_\mathcal{T}}

\newcommand{\centroidmat}{\mathcal{X}^{mat}_\mathcal{C}}
\newcommand{\centroid}{\mathcal{X}_\mathcal{C}}
\newcommand{\centroidfuture}{\mathcal{Y}_\mathcal{C}}
\newcommand{\centroidfuturemat}{\mathcal{Y}^{mat}_\mathcal{C}}
\newcommand{\setofinputsyn}{\mathcal{X}_\mathcal{S}}
\newcommand{\setoffuturesyn}{\mathcal{Y}_\mathcal{S}}
\newcommand{\series}{\mathcal{T}}
\newcommand{\seriessyn}{\mathcal{S}}
\newcommand{\seriessmall}{\mathcal{C}}
\newcommand{\augment}{DropLocation}

\definecolor{kindness}{RGB}{152, 78, 163}
\newcommand\yeongho[1]{\textcolor{kindness}{[Yeongho: #1]}}
\newtheorem{experiment}{\textbf{Preliminary Experiment}}
\newtheorem{problem}{\textbf{Problem}}
\newtheorem{thm}{\textbf{Theorem}}
\newtheorem{lma}{\textbf{Lemma}}

\SetKwComment{Comment}{$\triangleright$\ }{}
\title{Effective Dataset Distillation for Spatio-Temporal Forecasting with Bi-dimensional Compression
}

\author{
Taehyung Kwon\IEEEauthorrefmark{1},
Yeonje Choi\IEEEauthorrefmark{1},
Yeongho Kim,
Kijung Shin \\
\textit{Kim Jaechul Graduate School of AI, KAIST, Seoul, Republic of Korea} \\
\{taehyung.kwon, yeonjechoi, yeongho, kijungs\}@kaist.ac.kr
}

\maketitle
\begingroup\renewcommand\thefootnote{\IEEEauthorrefmark{1}}
\footnotetext{Equal contribution.}
\endgroup

\begin{abstract}
Spatio-temporal time series are widely used in real-world applications, including traffic prediction and weather forecasting.
They are sequences of observations over extensive periods and multiple locations, naturally represented as multi-dimensional data.
Forecasting is a central task in spatio-temporal analysis, and numerous deep learning methods have been developed to address it.
However, as dataset sizes and model complexities continue to grow in practice, training deep learning models has become increasingly time- and resource-intensive.


A promising solution to this challenge is dataset distillation, which synthesizes compact datasets that can effectively replace the original data for model training.
Although successful in various domains, including time series analysis, existing dataset-distillation methods compress only one dimension, making them less suitable for spatio-temporal datasets, where both spatial and temporal dimensions jointly contribute to the large data volume.





To address this limitation, we propose \method, the first dataset distillation method specialized for spatio-temporal time series forecasting.
A key idea of our solution is to compress both temporal and spatial dimensions in a balanced manner, reducing training time and memory.
We further reduce the distillation cost by performing distillation at the cluster level rather than the individual location level, and we complement this coarse-grained approach with a subset-based granular distillation technique that enhances forecasting performance.






On five real-world datasets, we show empirically that, compared to both general and time-series dataset distillation methods, datasets distilled by our \method method enable model training (1) \textbf{faster} (up to 6$\times$) (2) more \textbf{memory-efficient} (up to 8$\times$), and (3) more \textbf{effective} (with up to 12\% lower prediction error).
\end{abstract}


\section{Introduction}\label{intro}

\begin{figure*}[t]
    \vspace{-5mm}
    \centering
            \includegraphics[width=0.32\linewidth]{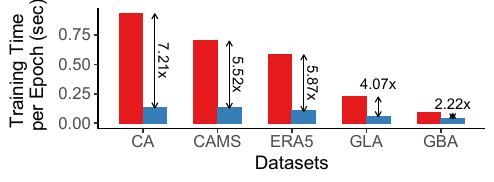}
            \includegraphics[width=0.32\linewidth]{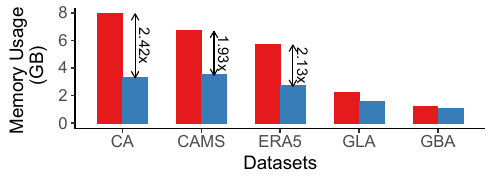}
        \includegraphics[width=0.1\linewidth]{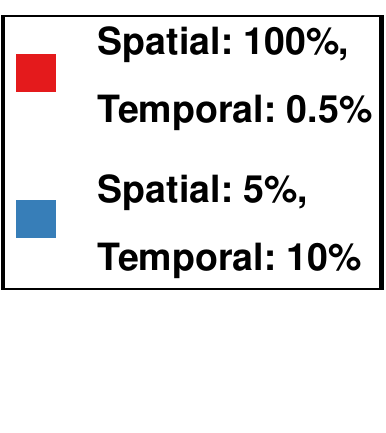}
    \vspace{-3mm}
    \caption{\label{fig:finding} Synthetic data compressed in both temporal and spatial dimensions in a balanced manner yields significantly lower training time and GPU memory usage compared to datasets of the same size compressed only along the temporal dimension. 
    See Preliminary Experiment~\ref{preliminary_experiment} for details. 
    }
\end{figure*}

Spatio-temporal time series datasets are universal, playing a key role in real-world applications, including traffic prediction~\cite{han2024bigst, fang2025efficient, cirstea2022towards, kieu2024team, lai2023lightcts} and weather forecasting~\cite{angryk2020multivariate, ma2023histgnn}. 
These datasets consist of observations from multiple locations over time, naturally represented as multi-dimensional arrays where different dimensions capture spatial and temporal information along with one or more measured features (e.g., wind, temperature, and pressure in weather datasets).
Since such time series are typically collected from numerous locations over extended periods~\cite{han2024bigst, fang2025efficient, miao2024unified}, they are often massive in scale.
A central task using spatio-temporal time series is forecasting~\cite{ma2023histgnn, cirstea2022towards, wu2024autocts++}, which aims to predict future time‑series values across locations based on past observations.
To this end, 
diverse deep learning methods~\cite{stgcn, wu2019graph, bai2020adaptive, wu2020connecting, shao2022decoupled} have been developed. 
However, as dataset sizes and model complexities continue to scale in real-world scenarios, training these models leads to significant storage demands and computation costs.


Dataset distillation has emerged as a promising approach to mitigate the heavy computational and storage costs of training deep learning models.
The key idea is to create a small and informative synthetic dataset that can replace original data in the training process while preserving comparable model performance. 
Several well-known approaches to it include gradient matching~\cite{zhao2021dataset}, trajectory matching~\cite{cazenavette2022dataset}, and 
meta-learning~\cite{wang2018dataset}.
Gradient matching minimizes the discrepancy between gradients computed on original and synthetic data, while trajectory matching aligns the training dynamics of models trained on both datasets.
Meta-learning focuses on improving the transferability of models trained on synthetic data to original datasets.
Although dataset distillation has been mainly studied for image and graph data~\cite{jin2021graph,jin2022condensing,gao2024graph, liang2025training}, recent work has explored its applications to time series~\cite{liu2024dataset, ding2024condtsf, miao2024less}.




However, since existing dataset distillation methods typically compress only a single dimension, they are less suitable for spatio-temporal datasets, where both spatial and temporal dimensions jointly contribute to the large data volume.
In particular, time-series distillation techniques compress only the temporal dimension, while leaving the spatial dimension (i.e., the number of locations) unchanged, which can lead to substantial training time and GPU memory usage, as evidenced by the preliminary experiment below.

\begin{experiment}[Necessity of Joint Reduction]
\label{preliminary_experiment}
In Figure~\ref{fig:finding}, we compare two random sampling strategies with \textbf{the same overall reduction ratio}: (1) reducing only the temporal dimension by 0.5\%, and (2) jointly reducing both dimensions by 5\% (spatial) and 10\% (temporal).
After applying the strategies to five real-world spatio-temporal time-series datasets (see Section~\ref{sec:exp:settings}), we measure the training time per outer iteration and GPU memory usage of MTGNN~\cite{mtgnn}, a representative deep learning model for spatio-temporal forecasting. 
Figure~\ref{fig:finding} shows that, even with the same overall reduction ratio, jointly reducing the spatial and temporal dimensions significantly reduces training time and GPU memory usage, compared to temporal-only reduction. Refer to Section~\ref{sec:exp:settings} for detailed experimental settings, and see also Theorems~\ref{thm:train:time} and \ref{thm:train:space} in Section~\ref{sec:method:complexity} for related analyses.
\end{experiment}

In this work, we propose \textbf{\method} (\textbf{S}patio-\textbf{Tem}peral Dataset \textbf{Dist}illation), the first dataset distillation method specialized for spatio-temporal time series. 
A crucial step needed to compress the spatial dimension is to design a surrogate model that can be trained on a smaller number of locations (i.e., spatially compressed synthetic data) yet remains applicable for inference on a larger number of locations, which is not supported by most existing spatio-temporal forecasting models.
To this end, we integrate a location encoder module that produces embeddings independent of the number of locations into Spatio-Temporal Graph Neural Networks (STGNNs), which are widely-used deep-learning models for spatio-temporal forecasting \cite{zhao2023multiple, ma2024learning, wu2021autocts, wu2024fully, cheng2023weakly, jiang2024sagdfn, cirstea2021enhancenet}. 

Using the surrogate model with the location encoder module, we build \method upon gradient matching (described above) with two enhancements. 
First, for efficiency, we reduce distillation time by performing distillation at the cluster level. Specifically, locations in the original dataset are grouped into clusters, and the averaged data within each cluster, rather than data from individual locations, are distilled. Second, to offset potential quality loss from this coarse-grained approach, we employ subset-based granular distillation. This technique distills data using varying subsets of clusters, ensuring that all parts (i.e., subsets) of the data are adequately reflected and thereby enhancing the quality of the output synthetic dataset.

Using five real-world spatio-temporal datasets and nine baseline approaches, we conduct a comprehensive evaluation of \method, which reveals the following advantages over the baselines under the same compression ratio:
\begin{itemize}[leftmargin=*]
    \item \textbf{Faster:} Model training is up to \textbf{$6\times$ faster} on synthetic data distilled by \method than on data distilled by the baselines.
    \item \textbf{Memory-efficient:} Model training requires up to \textbf{$8\times$ less GPU memory} on synthetic data distilled by \method than on data distilled by the baselines.
    \item \textbf{Effective:} Models trained on synthetic data distilled by \method achieve up to \textbf{12\% lower prediction error} compared to those trained on data distilled by the baselines.
\end{itemize}

The rest of the paper is organized as follows.
We present related works in Section \ref{related works}, preliminaries in Section \ref{Preliminaries}, and our proposed method in Section \ref{Method}.
We share experimental results in Section \ref{experiments} and conclude our paper in Section \ref{conclusion}.
\smallsection{Reproducibility}
The code and datasets are available at \url{https://github.com/kbrother/STemDist}.

\section{Related Works} \label{related works}

\smallsection{Spatio-temporal time series forecasting}
Spatio-temporal time series forecasting refers to the task of predicting future values of time series over a set of locations based on their historical observations~\cite{atluri2018spatio}. 
It is a key task in time series analysis that requires effectively capturing both temporal dynamics and spatial dependencies. 
Recent advances largely rely on Spatio-Temporal Graph Neural Networks
(STGNNs)~\cite{jin2023spatio, lan2022dstagnn,stnorm,shao2022spatial, kieu2024team}, which jointly model spatio-temporal dependencies and achieve state-of-the-art forecasting performance.
Previous works on STGNNs can be categorized into: 
(1) \textbf{Prior-based models}~\cite{fang2021spatial, fang2021mdtp, stgcn, li2017diffusion}, which model spatial dependencies between locations using a predefined adjacency matrix, i.e., explicit structural information, and
(2) \textbf{Structure-learning-based models}~\cite{shao2022pre, bai2020adaptive, wu2019graph, choi2022graph, mtgnn}, which infer the adjacency matrix end-to-end through a dedicated graph learning module. 

In this work, we adopt structure-learning-based models for dataset distillation due to their flexibility and scalability. Since the adjacency matrix is learned dynamically, neither predefined nor stored graph structures are needed during training or distillation. However, most STGNNs cannot be directly employed as surrogate models in our distillation framework, since they require the number of locations to be identical during training and testing, as elaborated in Section~\ref{sec:method:overview}.

As a particularly relevant STGNN-based approach, Zhou et al. \cite{zhou2025mogernn} proposed an inductive approach that infers embeddings for unobserved locations using an explicit graph structure and embeddings for observed locations. This is related to our location encoder (Section \ref{sec:method:node_model}), which also yields location embeddings. While their embeddings transfer knowledge from observed to unobserved locations, ours operate in the conceptually opposite direction by enabling knowledge transfer from a synthetic dataset to a real dataset without graph structure.

%


\smallsection{Dataset distillation for general data}
Dataset distillation aims to transform original datasets into smaller-scale synthetic datasets that serve as effective substitutes for the original datasets in machine-learning model training~\cite{sachdeva2023data}. Existing approaches can be categorized into:
(1) \textbf{Coreset selecting methods}~\cite{chai2023goodcore, welling2009herding, farahani2009facility, liang2025training, mirzasoleiman2020coresets}, which select a few representative samples from the original dataset,
(2) \textbf{Gradient matching based methods}~\cite{zhao2021dataset, gao2024graph,  zhao2021dataset2, wu2024condensing, gao2024heterogeneous}, which reduce the discrepancy between the gradients of models when trained on the original and synthetic datasets,
(3) \textbf{Trajectory matching based methods}~\cite{cazenavette2022dataset, guo2023towards}, which minimize the discrepancy between the training trajectories of model parameters obtained on the original and synthetic datasets, 
(4) \textbf{Meta-learning based methods}~\cite{zhou2022dataset}, 
which directly optimize the transferability of models trained on the synthetic dataset to the original dataset, and
(5) \textbf{Distribution matching based methods}~\cite{zhao2023dataset, zhao2023improved}, which directly align the output distributions of models initialized with random parameters. 

Our proposed distillation method, \method, builds upon gradient matching due to its simplicity and superior empirical effectiveness over alternatives on spatio-temporal time series datasets (refer to Section~\ref{sec:exp:main} for empirical comparisons).

\smallsection{Dataset distillation for time series}
Several dataset distillation methods have been developed specifically for time series datasets. CondTSF~\cite{ding2024condtsf} adapts general dataset distillation techniques to time series forecasting by incorporating a plugin that refines the synthetic dataset based on the outputs of a model trained on the original dataset. 
TimeDC~\cite{miao2024less} combines decomposition-driven frequency matching with trajectory matching. 
CondTSC~\cite{liu2024dataset} enhances the matching process through frequency-aware data augmentation, jointly leveraging temporal and frequency-domain information.

It is worth noting, however, that these methods are designed for uni- or multi-variate time series rather than spatio-temporal datasets, which are the focus of our work. Thus, when applied to spatio-temporal data, they compress only the temporal dimension while leaving the spatial dimension (i.e., the number of locations) unchanged, leading to substantial training costs even after distillation (see Preliminary Experiment~\ref{preliminary_experiment} in Section~\ref{intro}). Moreover, they are designed for and applied to models~\cite{wu2021autoformer, zhou2021informer} that do not explicitly capture correlations across different locations. In addition, CondTSC~\cite{liu2024dataset} is developed for classification, whereas our work targets forecasting.
These gaps motivate the development of dataset distillation methods specialized for spatio-temporal forecasting.

\smallsection{Dataset distillation for graphs}
Several studies~\cite{jin2021graph, jin2022condensing, zhang2024navigating, liu2022graph, wang2024fast} have explored graph dataset distillation.
They focus on reducing the number of nodes in a predefined graph topology.
Our problem differs from theirs in that spatio-temporal time series datasets lack a predefined graph structure, and we aim to compress both spatial and temporal dimensions.

\section{Preliminaries and problem definition}\label{Preliminaries}
\begin{table}[t]
\caption{Descriptions of frequently used notations} \label{tab:symbol}
\centering
\scalebox{0.95}{
     \begin{tabular}{c|l} 
         \toprule
         \textbf{Symbol} & \textbf{Description} \\          
         \midrule     
         $Z \in \mathbb{R}^{M \times F \times N}$ & Spatio-temporal time series \\
         $z_t \in \mathbb{R}^{F \times N} $ & $t$-th observation in $Z$ \\
         $M$ & Total length of time series \\
         $F$ & Number of (input) features \\         
         $N$ & Number of locations \\
         \midrule        
         $X_t \in \mathbb{R}^{L_{in} \times F \times N}$ & Input time series for time $t$  \\         
         $Y_t \in \mathbb{R}^{L_{out} \times F_{out} \times N}$ & Target time series for time $t$ \\
         $L_{in}$, $L_{out}$ & Lengths of input and target windows \\
         $F_{out}$ & Number of target features \\
         \midrule
         $\series = (\mathcal{X}_\series, \mathcal{Y}_\series)$ & Spatio-temporal time series dataset \\
         $\mathcal{X}_\mathcal{T} \in \mathbb{R}^{M_\series \times L_{in} \times F \times N_\series}$ & List of input time series in $\mathcal{T}$ \\
         $\mathcal{Y}_\mathcal{T} \in \mathbb{R}^{M_\series \times L_{out} \times F_{out} \times N_\series}$ & List of target time series in $\mathcal{T}$ \\
         $M_\series$ & Number of input-target pairs in $\series$ \\
         $N_\series$ & Number of locations in $\series$ \\
         \midrule
         $\seriessyn = (\mathcal{X}_\seriessyn, \mathcal{Y}_\seriessyn)$ & Synthetic time series dataset \\         
         $\mathcal{X}_\mathcal{S} \in \mathbb{R}^{M_\seriessyn \times L_{in} \times F \times N_\seriessyn}$ & List of input time series in $\mathcal{S}$ \\
         $\mathcal{Y}_\mathcal{S} \in \mathbb{R}^{M_\seriessyn \times L_{out} \times F_{out} \times N_\seriessyn}$ & List of target time series in $\seriessyn$ \\
         $M_\seriessyn$ & Number of input-target pairs in $\seriessyn$ \\
         $N_\seriessyn$ & Number of locations in $\seriessyn$ \\
         \midrule
         $\setofinputmat \in \mathbb{R}^{N_\series \times (M_{\series}L_{in}F)}$ & Reshaped version of $\setofinput$ \\
         $\seriessmall$ & Clustered time series dataset \\
         $K$ & Number of location sets\\
         $\{D_k\}_{k=0}^{K-1}$ & Divided location sets  \\
         \bottomrule 
     \end{tabular}}
\end{table}

This section presents the basic concepts and the formal definition of the problem, with a brief overview of gradient matching. 
Frequently used notations are listed in Table~\ref{tab:symbol}.

\subsection{Basic Concepts}
We first define a spatio-temporal time series as follows:
\begin{definition}[Spatio-temporal Time Series] \label{dfn:st_time_series}
A spatio-temporal time series $Z =(z_0, \cdots, z_{M-1})$ is a sequence of $M$ observations, where each observation $z_t \in \mathbb{R}^{F \times N}$ contains
$F$ features from $N$ spatial locations.
\end{definition}

Following prior works \cite{mtgnn, stgcn, bai2020adaptive, wu2019graph}, we define the spatio-temporal time series forecasting problem as predicting the future values of target features from a window of past observations, where the target features are typically a subset of the input features, as follows:
\begin{problem}\label{prob:forecasting}
\textsc{\normalfont{(Spatio-temporal Time Series Forecasting)}} 
\begin{itemize}[leftmargin=*]
    \item \textbf{Given:} a past input spatio-temporal time series $X_t = (z_{t-L_{in}}, \cdots,$ $z_{t-1})\in \mathbb{R}^{L_{in} \times F \times N}$ of length $L_{in}$,
    \item \textbf{Predict:} the future target spatio-temporal time series $Y_t=(z'_t, \cdots,$ $z'_{t + L_{out}-1})\in \mathbb{R}^{L_{out} \times F_{out} \times N}$ of length $L_{out}$,
\end{itemize}
    where $z'_t \in \mathbb{R}^{F_{out} \times N}$ is the $F_{out}$ target feature values at time~$t$.
\end{problem}





Using the input $X_t$ and target $Y_t$ defined above, we formally define the spatio-temporal time series dataset as follows.
\begin{definition}[Spatio-temporal Time Series Dataset] \label{dfn:st_time_dataset}
  A spatio-temporal time series dataset is defined as 
  \[
  \series = (\mathcal{X}_\series, \mathcal{Y}_\series) 
  = \big( (X_t)_{t=0}^{M_\series-1},\, (Y_t)_{t=0}^{M_\series-1} \big),
  \] 
  where $\mathcal{X}_\series=(X_t)_{t=0}^{M_\series-1}\in \mathbb{R}^{M_\series \times L_{in} \times F \times N_\series}$ denotes a list of input spatio-temporal time series, $\mathcal{Y}_\series=(Y_t)_{t=0}^{M_\series-1}\in \mathbb{R}^{M_\series \times L_{out} \times F_{out} \times N_\series}$ denotes a list of target spatio-temporal time series, $M_\series$ denotes the number of input and target spatio-temporal time-series pairs, and $N_\series$ denotes the number of locations in the dataset.
\end{definition}
\noindent
An example of creating the spatio-temporal time series dataset from spatio-temporal time series is in Appendix A~\cite{appendix}.



Similarly, we define a synthetic spatio-temporal time series dataset (hereafter referred to as a synthetic dataset), which is obtained through dataset distillation, as follows: 
\begin{definition}[Synthetic Dataset] \label{dfn:syn_dataset}
  A synthetic spatio-temporal time series dataset is defined as 
  \[
  \mathcal{S} = (\mathcal{X}_\seriessyn, \mathcal{Y}_\seriessyn) 
  = \big((\tilde{X}_i)_{i=0}^{M_\seriessyn - 1}, (\tilde{Y}_i)_{i=0}^{M_\seriessyn - 1}\big),
  \] 
  where $\mathcal{X}_\seriessyn=(\tilde{X}_i)_{i=0}^{M_\seriessyn-1}\in \mathbb{R}^{M_\seriessyn \times L_{in} \times F \times N_\seriessyn}$ denotes a list of input synthetic time series, $\mathcal{Y}_\seriessyn=(\tilde{Y}_i)_{i=0}^{M_\seriessyn - 1}\in\mathbb{R}^{M_\seriessyn \times L_{out} \times F_{out} \times N_\seriessyn}$ denotes a list of target synthetic time series, $M_\seriessyn$ denotes the number of input-target pairs, and $N_\seriessyn$ denotes the number of locations in the synthetic dataset.
  
\end{definition}

\subsection{Problem Definition}
Building on the basic concepts, we now define the problem of dataset distillation for spatio-temporal time series forecasting.
Let $g_\theta$ denote a (predefined or arbitrary) machine learning model for spatio-temporal time series forecasting with parameters $\theta$; 
and $\theta_\seriessyn$ is the set of parameters obtained by training on $\seriessyn$.
Dataset distillation aims to construct a synthetic dataset $\seriessyn$ such that the model trained on it, $g_{\theta_\seriessyn}$, achieves the best performance on ground-truth data distribution $P$, i.e.,
\begin{equation}
    \min_{\mathcal{S}}{
    \mathbb{E}_{(X,Y) \sim P}[\ell (g_{\theta_\seriessyn}(X), Y)]},
\end{equation} 
where $\ell$ is a loss function  (e.g., sum of squared error).
Since the ground-truth data distribution $P$ is unknown in practice, we formulate the problem using a given dataset $\mathcal{T}$ as follows:
\begin{problem} \label{prob:condensation}
\textsc{\normalfont{(Dataset Distillation for Spatio-temporal Time Series Forecasting)}}
    \begin{itemize}[leftmargin=*]
        \item \textbf{Given:} a spatio-temporal time series dataset $\mathcal{T}$,
        \item \textbf{Find:} a synthetic dataset $\mathcal{S}$,
        \item \textbf{to Minimize:} the forecasting error of $g_{\theta_S}$ on $\mathcal{T}$, i.e.,
        \[
        \min\nolimits_{\mathcal{S}} \mathcal{L}(g_{\theta_{\mathcal{S}}}, \mathcal{T}) \quad \text{subject to} \quad \theta_{\mathcal{S}} = \arg\min\nolimits_{\theta} \mathcal{L}(g_\theta, \mathcal{S}),
        \]
    \end{itemize}
where $\mathcal{L}(g_{\theta_S}, \mathcal{T})=\frac{1}{M_\series}\sum_{t=0}^{M_\series - 1} \ell(g_{\theta_S}(X_t), Y_t)$ is the average prediction error of $g_{\theta_S}$ on $\mathcal{T}$; and $\mathcal{L}(g_\theta, \mathcal{S})=\frac{1}{M_\seriessyn} \sum_{i=0}^{M_\seriessyn - 1}\ell(g_\theta(\tilde{X}_i), \tilde{Y}_i)$ is that of $g_{\theta}$ on~$\mathcal{S}$. 
\end{problem}
\vspace{-1mm}



\subsection{Gradient Matching}
\label{sec:gradient_matching}

In this section, we briefly review gradient matching~\cite{zhao2021dataset}, a dataset distillation method that our method builds upon.

Gradient matching aims to learn synthetic data such that the model parameters $\theta_\seriessyn$ trained on it converge to a solution close to parameters $\theta_\series$ obtained from model training on original data.
However, 
$\theta_\seriessyn$ often converges to a local minimum different from $\theta_\series$, making direct alignment with $\theta_\series$ ineffective.
Gradient matching addresses this issue by aligning the gradients computed on $\series$ and $\seriessyn$ at every update of $\theta_\seriessyn$, even before convergence, without depending on the values of $\theta_\series$.

To formalize the gradient matching objective, we denote the parameters of a surrogate machine learning model $f_\theta$ after $t$ optimization steps as $\theta_t$, initialized with $\theta_0$.
The objective is to minimize the distance between the gradients computed on synthetic data $\seriessyn$ and original data $\series$ along the optimization trajectory from $t=0$ to $T_{max}-1$, i.e.,
\vspace{-1mm}
\[\min_{\mathcal{S}} \mathbb{E}_{\theta_0 \sim P_{\theta_0}} \left[\sum_{t=0}^{T_{max}-1} Dist(\nabla_\theta \mathcal{L}(\seriessyn, \theta_t), \nabla_\theta \mathcal{L}(\series, \theta_t)) \right], \]
\[\text{subject to} \quad\theta_t \leftarrow \theta_{t-1} - \eta \nabla_\theta \mathcal{L}(S, \theta_{t-1}),   \]
where $P_{\theta_0}$ is the distribution of the initial parameter, $\eta$ is the learning rate, $Dist$ is the sum of cosine distances between two gradients, and $\mathcal{L}$ is the training loss function.
\vspace{-1mm}


\section{Proposed Method}\label{Method}

In this section, we present \method (\textbf{S}patio-\textbf{Tem}peral Dataset \textbf{Dist}illation), our proposed dataset distillation method for spatio-temporal time-series forecasting. We first outline the key challenges in solving Problem~\ref{prob:condensation} and our corresponding solutions (Section~\ref{sec:method:overview}).
Then, we provide an overview of \method (Section~\ref{sec:method:distillation}) and describe the three key components of \method: location encoders (Section~\ref{sec:method:node_model}), clustering (Section~\ref{sec:method:clustering}), and granular distillation (Section~\ref{sec:method:augmentation}). Lastly, we provide a complexity analysis (Section~\ref{sec:method:complexity}).

\subsection{Key Challenges and Our Corresponding Solutions}
\label{sec:method:overview}
There are three key challenges in solving Problem~\ref{prob:condensation}:
\begin{itemize}[leftmargin=*]


\item \textbf{C1. Existing methods insufficiently reduce training cost.} 
Existing dataset distillation methods focus on compressing only a single dimension, the temporal dimension in our context (i.e., $M_\seriessyn$ in Definition~\ref{dfn:syn_dataset}). However, the training cost of forecasting models is also affected by the number of locations (i.e., $N_\seriessyn$ in Definition~\ref{dfn:syn_dataset}), which contributes quadratically to the computational cost. Thus, training remains expensive without compressing the spatial dimension (i.e., the number of locations).  

\item \textbf{C2: Dataset distillation itself is time-consuming.} 
The ultimate goal of dataset distillation is to reduce the overall computational cost associated with model training.
Thus, if a distillation process itself is too costly, then regardless of its performance, a dataset distillation method fails to serve its intended purpose.
Therefore, the distillation cost, which often grows substantially with the number of locations (i.e., $N_\series$ in Definition~\ref{dfn:st_time_dataset}), must remain computationally efficient.


\item \textbf{C3: Synthetic datasets may not fully capture original data.}
Real-world spatio-temporal time series are often collected from a large number of locations.
The data from every location should be effectively distilled for effective model training on the synthetic data.
However, when the number of locations is large, it becomes challenging to account for all of them during distillation, particularly given the limited capacity of synthetic datasets.





\end{itemize}

To tackle these challenges, our proposed method, \method, introduces three corresponding solutions:
\begin{itemize}[leftmargin=*]
\item \textbf{S1. Simultaneous compression of temporal and spatial dimensions with location encoders.} 
To address C1, we jointly compress both temporal and spatial dimensions (i.e., $M_\seriessyn$ and $N_\seriessyn$ in Definition~\ref{dfn:syn_dataset}). 
To the best of our knowledge, we are the first to propose such a bi-dimensional dataset distillation strategy. A key to implementing this idea is the introduction of location encoders (see Section~\ref{sec:method:node_model}), which allow deep learning models trained on a smaller number of locations to be applied for inference on a larger number of locations.
Jointly compressing both dimensions yields a synthetic dataset that balances both aspects, which not only lowers training cost but also improves model performance.


\item \textbf{S2. Clustering of locations in original datasets.}
To tackle C2, we reduce the number of locations in the original dataset (i.e., $N_\series$ in Definition~\ref{dfn:st_time_dataset}) by clustering them. The time series of individual locations are replaced with those of the cluster centroids, resulting in a reduced version of the original dataset, as detailed in Section~\ref{sec:method:clustering}.
Distilling this clustered dataset speeds up the overall distillation process.
Note that this idea of reducing the number of locations in the original dataset ($N_\series$) is distinct from the earlier idea of reducing that in synthetic datasets ($N_\seriessyn$).

\item \textbf{S3. Subset-based granular distillation.} 
To address C3, we propose subset-based granular distillation. Instead of using all locations at once, the distillation process leverages varying subsets of locations. This allows different parts (i.e., subsets) of the data to be more effectively reflected in the synthetic dataset, improving its overall quality.
Note that this idea complements the cluster-based coarse-grained approach by mitigating potential quality loss caused by clustering.
\end{itemize}


These solutions reflect the central idea of our method, and their details  are provided in Sections \ref{sec:method:node_model}, \ref{sec:method:clustering}, and \ref{sec:method:augmentation}.

\subsection{Overview of \method} \label{sec:method:distillation}

In this subsection, we provide an overview of \method, with its overall procedure summarized in Algorithm~\ref{algo:distill}.

 


Lines \ref{algo:distill:cluster} - \ref{algo:distill:init} describe the preprocessing steps before distillation.
First, locations in the original dataset $\series$ are clustered to obtain a spatially reduced version, $\seriessmall=(\centroid, \centroidfuture)$, and the cluster weights 
$(w_i)_{i=0}^{N_\seriessyn - 1}$, proportional to the number of locations in each cluster
(line~\ref{algo:distill:cluster}).
The clustered dataset $\seriessmall$ replaces $\series$ in subsequent steps, accelerating the distillation process (see \textbf{S2} in Section~\ref{sec:method:overview}).
Details for the clustering process are provided in Section~\ref{sec:method:clustering} (see Algorithm~\ref{algo:clustering}).
Then, the input of the location encoders (see \textbf{S1} in Section~\ref{sec:method:overview}), $I_\seriessmall$ is computed from $\mathcal{X}_{\seriessmall}$ in advance (line \ref{algo:distill:input_real}) and subsequently used in the following steps.
The synthetic dataset $\seriessyn=(\setofinputsyn, \setoffuturesyn)$ is initialized by random sampling from $\seriessmall$ (line \ref{algo:distill:init}), naturally assigning each cluster to a synthetic location in $\seriessyn$.\footnote{
As distillation progresses, the semantic meaning of synthetic locations may be obscured, since distillation is designed to optimize training equivalence rather than interpretability. Thus, synthetic locations should be interpreted as training-efficient proxies rather than semantically meaningful locations.
}

Lines \ref{algo:distill:init_param} - \ref{algo:distill:model_train} describe the distillation process.
Given the preprocessed inputs ($\seriessmall$, $I_\seriessmall$, and initialized $\seriessyn$), it outputs the distilled synthetic data $\seriessyn$.
Since the clustered dataset $\seriessmall$ can be large along the temporal dimension, we perform distillation using mini-batch pairs $(\centroid^{B}, \centroidfuture^{B})$ sampled along the temporal dimension from $\seriessmall=(\centroid, \centroidfuture)$ (line \ref{algo:distill:mini_batch}).
For subset-based granular distillation (see \textbf{S3} in Section~\ref{sec:method:overview}), we partition the locations into $K$ disjoint subsets $\{D_k\}_{k=0}^{K-1}$ (line \ref{algo:distill:partition}).

For each location subset $D_k$, we perform gradient matching (see Section~\ref{sec:gradient_matching}) to update the synthetic time series of the corresponding synthetic locations (lines \ref{algo:distill:input_syn} - \ref{algo:distill:backprop}), and employ 
an STGNN equipped with our location encoder module (see Section \ref{sec:method:node_model}) as 
a surrogate model $f_\theta$.
A mini-batch $\centroid^{B}$ of the input time series from the clustered dataset $\seriessmall$ and the input time series $\setofinputsyn$ of the synthetic dataset $\seriessyn$ are given to the model $f_\theta$ along with the inputs of the location encoders, $I_\seriessmall$ and $I_\seriessyn$ (lines \ref{algo:distill:output_real} and \ref{algo:distill:output_syn}). 
Then, the losses for $\seriessmall$ and $\seriessyn$
are computed
considering the weights $(w_i)_{i=0}^{N_\seriessyn - 1}$ of clusters, assigned to each synthetic location (lines \ref{algo:distill:loss_real} and \ref{algo:distill:loss_syn}). 
After that, the discrepancy between the gradients is measured, yielding the gradient matching loss (line~\ref{algo:distill:grad_match}), which is backpropagated to update the synthetic dataset $\seriessyn$
(line~\ref{algo:distill:backprop}).
After updating the synthetic dataset for all location subsets, the parameters of $f_\theta$ are updated using the current synthetic dataset (line \ref{algo:distill:model_train}).

The detailed training process of the surrogate model $f_\theta$ on the synthetic dataset $\seriessyn=(\setofinputsyn, \setoffuturesyn)$ is provided in Algorithm~\ref{algo:train}.
Given $\seriessyn$, we first compute the input $I_\seriessyn$ for the location encoders (line \ref{algo:train:input}).
For each training epoch $t$, the locations in the synthetic dataset are randomly partitioned into $K$ disjoint subsets (line \ref{algo:train:partition}), to be consistent with the subset-level granularity used in the distillation stage.
For each subset $D_k$, the model $f_\theta$ generates predictions for the corresponding inputs of the synthetic dataset (line \ref{algo:train:pred}).
The weighted loss used also in the distillation process is computed (line~\ref{algo:train:loss}), and finally, the model parameters are updated by gradient descent (line~\ref{algo:train:gd}).

\begin{algorithm}[t]
\small
\caption{Dataset Distillation in \method} \label{algo:distill}
\SetKwInput{KwInput}{Input} 
\SetKwInput{KwOutput}{Output}
\KwInput{(1) the original dataset $\series=(\setofinput, \setoffuture)$, \\ \hspace{8.5mm} (2) a number $T_{outer}$ of outer iterations, \\ \hspace{8.5mm} (3) a number $T_{distill}$ of iterations for distillation, \\ \hspace{8.5mm}  (4) a number $T_{model}$ of epochs for model training, \\ \hspace{8.5mm}  (5) a number $K$ of location sets, \\ \hspace{8.5mm}  (6) a learning rate $\eta$ for $\theta_{t}$.}
\KwOutput{Synthetic dataset $\seriessyn=(\setofinputsyn, \setoffuturesyn)$.}



        

Cluster the locations in $\series$ to get the clustered dataset $\seriessmall=(\centroid, \centroidfuture)$ and cluster weights $(w_i)_{i=0}^{N_\seriessyn - 1}$ \label{algo:distill:cluster} \\ \blue{\Comment*[f]{Algo. \ref{algo:clustering} and Sect. \ref{sec:method:clustering}}} 

Compute the input $I_\seriessmall$ of the location encoder from $\centroid$ \label{algo:distill:input_real} \\ \blue{\Comment*[f]{Sect. \ref{sec:method:node_model}}} 

Initialize $\seriessyn=(\setofinputsyn, \setoffuturesyn)$ by random sampling from $(\centroid, \centroidfuture)$ \label{algo:distill:init}

\For{$t_{outer} \leftarrow 0, \cdots, T_{outer}-1$}
{

    Initialize the surrogate model $f_\theta$ with $\theta_0 \sim P_{\theta_0}$  \label{algo:distill:init_param}
    
    \For{$t \leftarrow 0, \cdots, T_{distill}-1$}
    {
        Sample a mini-batch pair $\centroid^{B} \sim \centroid$ and $\centroidfuture^{B} \sim \centroidfuture$  \label{algo:distill:mini_batch}

        Partition locations $\{0, \cdots, N_\seriessyn-1\}$ into $\{D_k\}_{k=0}^{K-1}$ \label{algo:distill:partition} \\ \blue{\Comment*[f]{Sect. \ref{sec:method:augmentation}}}

        \For{$k \leftarrow 0, \cdots, K-1$}
        {

            
            Compute the input $I_\seriessyn$ of the location encoder from $\mathcal{X}_{\seriessyn}$ \label{algo:distill:input_syn}  \blue{\Comment*[f]{Sect. \ref{sec:method:node_model}}} 
            
            $\Tilde{\mathcal{Y}}_{\seriessmall} \leftarrow f_{\theta_t}(\centroid^{B}(:,:,:,D_k), I_\seriessmall(D_k,:))$  \label{algo:distill:output_real}


            $\mathcal{L}(\seriessmall, \theta_t) \leftarrow \sum_{i \in D_k} w_i \| (\Tilde{\mathcal{Y}}_{\seriessmall}- \mathcal{Y}_{\seriessmall}^B) (:,:,:,i)\|_2^2$
            \label{algo:distill:loss_real}
            
            $\Tilde{\mathcal{Y}}_{\seriessyn} \leftarrow f_{\theta_t}(\setofinputsyn(:,:,:,D_k), I_\seriessyn(D_k,:))$  \label{algo:distill:output_syn} 

            $\mathcal{L}(\seriessyn, \theta_t) \leftarrow \sum_{i \in D_k} w_i \| (\Tilde{\mathcal{Y}}_{\seriessyn} - \mathcal{Y}_{\seriessyn}) (:,:,:,i)\|_2^2$ \label{algo:distill:loss_syn}

            $\mathcal{L}^{grad} \leftarrow Dist(\nabla_\theta \mathcal{L}(\seriessmall, \theta_t), \nabla_\theta \mathcal{L}(\seriessyn, \theta_t))$  \label{algo:distill:grad_match}

            Backpropagate $\mathcal{L}^{grad}$ and update $\setofinputsyn(:,:,:,D_k)$ and $\setoffuturesyn(:,:,:,D_k)$   \label{algo:distill:backprop} \blue{\Comment*[f]{Sect. \ref{sec:method:augmentation}}}
        }
        


    $\theta_{t+1} \leftarrow$ Training the surrogate model $f_\theta$ using $(\setofinputsyn,$ $\setoffuturesyn, T_{model}, K, \eta,\theta_t,$ and $(w_i)_{i=0}^{N_\seriessyn - 1})$ \label{algo:distill:model_train} \blue{\Comment*[f]{Algo. \ref{algo:train}}}   
        
    }
}
\Return $(\setofinputsyn$, $\setoffuturesyn)$
\end{algorithm}
\vspace{-2mm}
\begin{algorithm}[t]
\small
\caption{(Surrogate) Model Training in \method} \label{algo:train}
\SetKwInput{KwInput}{Input} 
\SetKwInput{KwOutput}{Output}
\KwInput{(1) a synthetic dataset $\seriessyn=(\setofinputsyn, \setoffuturesyn)$, \\  \hspace{8.5mm} (2) a number $T$ of training epochs,  \\ \hspace{8.5mm} (3) a number $K$ of locations sets,  \\ \hspace{8.5mm} (4) a learning rate $\eta$, initial parameters $\theta_0$,  \\ \hspace{8.5mm} (5) cluster weights $(w_i)_{i=0}^{N_\seriessyn - 1}$.}
\KwOutput{trained parameters $\theta_T$ of the (surrogate) model $f_\theta$.}
Compute the input $I_\seriessyn$ of the location encoder from $\mathcal{X}_{\seriessyn}$  \label{algo:train:input} \\ \blue{\Comment*[f]{Sect.~\ref{sec:method:node_model}}}         
        
\For{$t \leftarrow 0,\cdots, T-1$} 
{   
    $\{D_k\}_{k=0}^{K-1} \leftarrow$ Partition locations $\{0, \cdots, N_\seriessyn-1\}$ \label{algo:train:partition} \\ \blue{\Comment*[f]{Sect. \ref{sec:method:augmentation}}}   

    \For{$k \leftarrow 0, \cdots, K-1$}
    {
        $\Tilde{\mathcal{Y}}_{\seriessyn} \leftarrow f_{\theta_{t}}(\setofinputsyn(:,:,:,D_k), I_\seriessyn(D_k,:))$ \label{algo:train:pred}

        
        $\mathcal{L}(\seriessyn, \theta_{t}) \leftarrow \sum_{i \in D_k} w_i \| (\Tilde{\mathcal{Y}}_{\seriessyn} - \mathcal{Y}_{\seriessyn}) (:,:,:,i)\|_2^2$ \label{algo:train:loss}

        $\theta_{t} \leftarrow \theta_{t} - \eta \nabla_\theta \mathcal{L}(\seriessyn, \theta_{t}) $   \label{algo:train:gd}
    } 

    $\theta_{t+1} \leftarrow \theta_t$
}

\Return $\theta_T$
\end{algorithm}

\subsection{Location Encoder} 
\label{sec:method:node_model}

In this subsection, we elaborate on our first solution, \textbf{S1} introduced in Section~\ref{sec:method:overview}. 
Specifically, we present the location encoder module that enhances STGNNs (described in Section~\ref{related works}), which serve as the surrogate model $f_\theta$ in the distillation process and, more generally, as powerful models for spatio-temporal time series forecasting.

As described in \textbf{C1} in Section~\ref{sec:method:overview}, while compressing the spatial dimension (in addition to the temporal dimension) is desirable for dataset distillation, it is not straightforward because STGNNs (and also most alternatives) trained
on a (synthetic) dataset with fewer locations are inapplicable for inference on a larger number of locations.
To address this, we introduce a location encoder that makes STGNNs inductive with respect to the number of locations, as described below.

\smallsection{Transductivity of STGNNs}
We first describe the components of STGNNs that make them inapplicable when the locations differ between training and inference.
In the absence of a predefined graph (i.e., spatial dependencies between locations), STGNNs~\cite{bai2020adaptive, wu2019graph, mtgnn} learn the graph structure through transductive (i.e., directly learnable) location embeddings.
For example, in AGCRN~\cite{bai2020adaptive} and graph wavenet~\cite{wu2019graph}, an adjacency matrix $A$ is derived as follows:
\begin{equation} \label{eq:adj}
    A \leftarrow \mathrm{Softmax}(\mathrm{ReLU}(E_1^T E_2)),
\end{equation}
where $E_1 \in \mathbb{R}^{N \times R}$ and $E_2 \in \mathbb{R}^{N \times R}$ are transductive embeddings of dimension $R$ for $N$ locations, and $\mathrm{Softmax}$ and $\mathrm{ReLU}$ are activation functions.
Note that the trained parameters are directly tied to the locations in the training dataset (i.e., the synthetic dataset in our dataset distillation context), making the STGNNs transductive. That is, when the semantics of locations change or when new locations are introduced during inference, the learned embeddings and the derived adjacency matrix become inapplicable.


\smallsection{Overview of location encoders}
Our approach to addressing this limitation is to use a location encoder that can replace the transductive location embeddings.
Specifically, for spatio-temporal time series with an arbitrary number of locations, the encoder is designed to generate embeddings for each location in the dataset. 
Especially, even when trained with a smaller number of (synthetic) locations in the synthetic dataset $\seriessyn$, the encoder should be capable of generalizing to a larger number of locations in the original dataset $\series$ during inference.

To this end, we design the location encoder as a sequence-to-sequence architecture, since both the inputs (time series from multiple locations) and the outputs (corresponding location embeddings) are naturally represented as sequences indexed by locations.
The encoder is further required to satisfy \textbf{(1) Length-insensitivity:} it should process sequences of arbitrary length, enabling it to handle both the synthetic and original datasets with different numbers of locations, and \textbf{(2) Parameter sharing:} it uses shared parameters to generate each location embedding, facilitating consistency and generalization 
across varying numbers of locations.
Note that, while we focus on a sequence-to-sequence design, other architectures satisfying these properties may also be viable.


\smallsection{Inputs and outputs of location encoders}
To obtain the input to the location encoder, we preprocess $\setofinput \in \mathbb{R}^{M_\series \times L_{in} \times F \times N_\series}$ when using the original data $\series$ and $\setofinputsyn \in \mathbb{R}^{M_\seriessyn \times L_{in} \times F \times N_\seriessyn}$ when using the synthetic data $\seriessyn$ (refer to Definitions \ref{dfn:st_time_dataset} and \ref{dfn:syn_dataset} for $\setofinput$ and $\setofinputsyn$).
Since they can be excessively long, we first average them along the temporal dimension. 
That is, $\bar{\mathcal{X}}_\series$ and $\bar{\mathcal{X}}_\seriessyn$ are computed as follows:
\begin{equation}
     \bar{\mathcal{X}}_{(\series, \seriessyn)} \leftarrow \frac{1}{M_{(\series, \seriessyn)}} \sum\nolimits_{m=0}^{M_{(\series, \seriessyn)} - 1} \mathcal{X}_{(\series, \seriessyn)}(m, :, :, :).
\label{equ:location}
\vspace{-1mm}
\end{equation}
Then, we reshape $\bar{\mathcal{X}}_\series \in \mathbb{R}^{L_{in} \times F \times N_\series}$ and $\bar{\mathcal{X}}_\seriessyn \in \mathbb{R}^{L_{in} \times F \times N_\seriessyn}$ into matrices $I_\series \in \mathbb{R}^{N_\series \times L_{in}F}$ and $I_\seriessyn \in \mathbb{R}^{N_\seriessyn \times L_{in}F}$, respectively, which are used as the input of the location encoder.

Given the input $I_\series$ (or $I_\seriessyn$), the location encoder outputs a matrix $E_\series \in \mathbb{R}^{N_\series \times R}$ (or $E_\seriessyn \in \mathbb{R}^{N_\seriessyn \times R}$), where each row represents an $R$-dimensional embedding of the corresponding location.
These output matrices are used (in place of the transductive location embeddings) to construct the adjacency matrix, following the structure-learning paradigm of STGNNs.

\blue{

}


\begin{figure*}[t]
    \vspace{-5mm}
    \centering
    \includegraphics[width=0.90\linewidth]{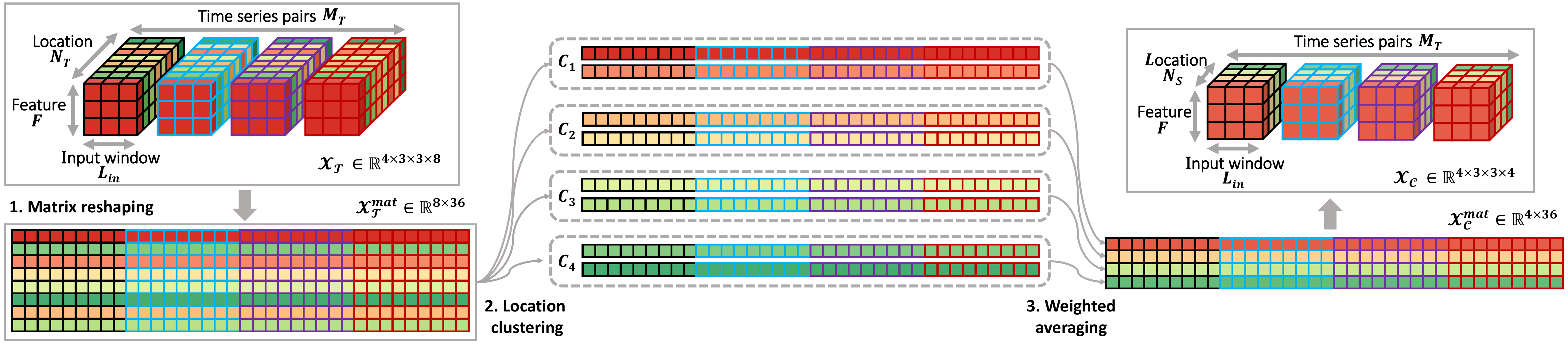}
    \vspace{-2mm}
    \caption{An example illustrating location clustering, where eight locations in $\setofinput$ are reduced into four locations (cluster centroids) in $\centroid$. Entries with the same color correspond to the data for the same location. Refer to Example~\ref{cluster:example} for details.}
    \label{fig:example}
\end{figure*}

\smallsection{Detailed architecture of location encoders}
The location encoder consists of a single-head self-attention layer~\cite{vaswani2017attention} preceded and followed by linear layers. 
A residual connection is applied around the attention layer, followed by layer normalization~\cite{ba2016layer}.
For clarity, we present the equations using $I_\series$, but they can also be expressed with $I_\seriessyn$.
\begin{equation} \label{eq:location_encoder}
\begin{split}
& H_1  = \mathrm{Linear_1}(I_\series), ~ H_2 = \mathrm{SelfAttn}(H_1), \\
& E_\series = \mathrm{Linear_2}\big(\mathrm{LayerNorm}(H_1 + H_2)\big).
\end{split}
\end{equation}
The hidden vectors are denoted by $H_1 \in \mathbb{R}^{N_\series \times d}$ and $H_2 \in \mathbb{R}^{N_\series \times d}$,  with $d$ representing the hidden dimension.
By the self-attention layer ($\mathrm{SelfAttn}$), 
each $i$-th row of $H_2$, the hidden vector for the $i$-th location, is computed as follows:
\begin{align} \label{eq:attn}
    & H_2(i,:)   = \sum\nolimits_{j=1}^{N_\series} \alpha(i,j) H_1(j,:) W_V W_O, \text{  where}\\         
    \alpha(i,j) &  = \frac{\mathrm{exp}\big((H_1(i,:) W_Q) \cdot (H_1(j,:) W_K)^T / \sqrt{d} \big)}{\sum_{n=1}^{N_\series} \mathrm{exp}\big((H_1(i,:) W_Q) \cdot (H_1(n,:) W_K)^T / \sqrt{d} \big)}. \nonumber
\end{align}
Note that $W_Q,W_K,W_V$, and $W_O \in \mathbb{R}^{d \times d}$ are learnable parameters.
The self-attention layer computes the importance score $\alpha(i,j)$ of each $j$-th location with respect to the $i$-th location. 
Then, the information extracted from each $j$-th location, $H_1(j,:)W_V$, is aggregated using $\alpha(i,j)$ as the weight to produce the hidden vector for the $i$-th location, $H_2(i,:)$.
Thus, the resulting hidden vector $H_2(i,:)$ captures the information from all locations, weighted by their learned relevance to the $i$-th location.

This architecture satisfies the two aforementioned properties required by location encoders.
All components in Eq.~\eqref{eq:location_encoder} and Eq.~\eqref{eq:attn} are \textbf{length-insensitive}, processing sequences of arbitrary length with parameter sizes independent of the sequence length.
In addition, all parameters (i.e., those of the linear layers, layer normalization, and $W_Q$, $W_K$, $W_V$, and $W_O$) are shared across all locations.
Thus, \textbf{parameter sharing} is achieved facilitating consistency and generalization across varying numbers of locations.
Therefore, our location encoder enables STGNNs to be applicable and effective for datasets with varying numbers of locations, including the original dataset with more locations than the synthetic training dataset.

\begin{algorithm}[t]
\small
\caption{Location Clustering in \method} \label{algo:clustering}
\SetKwInput{KwInput}{Input} 
\SetKwInput{KwOutput}{Output}
\KwInput{(1) the original dataset $(\setofinput, \setoffuture)$
\\ \hspace{8.5mm} (2) a number $T_{kmeans}$ of K-means iterations
}
\KwOutput{(1) the clustered dataset $(\centroid, \centroidfuture)$ \\ \hspace{10.7mm} (2) the cluster weights $(w_i)_{i=0}^{N_\seriessyn - 1}$}
Reshape $\setofinput$ to $\setofinputmat$ and $\setoffuture$ to $\setoffuturemat$ \label{algo:clustering:reshape}

$\{C_i\}_{i=0}^{N_\seriessyn - 1} \leftarrow$ K-means($\setofinputmat$, $T_{kmeans}$)  \label{algo:clustering:kmeans}

\For{$i \leftarrow 0$ to $N_\seriessyn - 1$}
{

        $(\centroidmat, \centroidfuturemat)(i,:) \leftarrow \frac{1}{|C_i|} \sum_{l \in C_i} (\setofinputmat, \setoffuturemat)(l, :)$  \label{algo:clustering:x}
        

        $w_i \leftarrow \dfrac{|C_i|}{N_\series}$   \label{algo:clustering:weight}
}
Reshape $\centroidmat$ to $\centroid$ and $\centroidfuturemat$ to $\centroidfuture$

\Return $(\centroid, \centroidfuture)$, $(w_i)_{i=0}^{N_\seriessyn - 1}$
\end{algorithm} 

\subsection{Clustering of Locations} \label{sec:method:clustering}
As discussed in \textbf{C2} in Section~\ref{sec:method:overview}, reducing the number of locations in the original dataset is necessary for efficient distillation. 
We address this by location clustering (\textbf{A2}) described in Algorithm~\ref{algo:clustering}, which consists of three steps:
\begin{enumerate}[leftmargin=*]
    \item We reshape the input-target time series pairs $(\setofinput, \setoffuture)$ $\in$ $(\mathbb{R}^{M_\series \times L_{in} \times F \times N_\series}, \mathbb{R}^{M_\series \times L_{out} \times F_{out} \times N_\series})$  of the original dataset $\series$ into matrices $(\setofinputmat, \setoffuturemat) \in (\mathbb{R}^{N_\series \times (M_\series L_{in} F)}, \mathbb{R}^{N_\series \times (M_\series L_{out} F_{out})})$ (line \ref{algo:clustering:reshape}). Each row of $\setofinputmat$, $\setoffuturemat$ corresponds to a location in $\series$.
    
    \item Then, we treat each row of $\setofinputmat$ as a feature vector for a location and apply K-means clustering on  $\setofinputmat$ to group $N_\series$ locations into $N_\seriessyn$ clusters (line \ref{algo:clustering:kmeans}). For each cluster $C_i$, we compute its representative feature vector by averaging over all the feature vectors of the locations assigned to $C_i$, yielding $\centroidmat(i,:)$ and $\centroidfuturemat(i,:)$ (lines \ref{algo:clustering:x}).
    
    \item Lastly, we reshape the obtained features of the clusters $(\centroidmat, \centroidfuturemat)\in(\mathbb{R}^{N_\seriessyn \times (M_\series L_{in} F)}, $
    $\mathbb{R}^{N_\seriessyn \times (M_\series L_{out} F_{out})})$ into original dimensions $(\centroid,\centroidfuture) \in $ $(\mathbb{R}^{M_\series \times L_{in} \times F \times N_\seriessyn},$ $\mathbb{R}^{M_\series \times L_{out} \times F_{out} \times N_\seriessyn})$, resulting in the clustered dataset $\seriessmall$.
    We treat $(\centroid, \centroidfuture)$ as the input–target time series pairs from the reduced locations, each representing a cluster.
\end{enumerate}

\begin{example} \label{cluster:example}
    In Figure~\ref{fig:example}, we illustrate the clustering process for the input time series $\setofinput$.
    The same procedure is applied to the target time series $\setoffuture$ to obtain $\centroidfuture$.
    Entries with the same color correspond to the same location.
    First, $\setofinput \in  \mathbb{R}^{4 \times 3 \times 3 \times 8}$ is reshaped into $\setofinputmat \in \mathbb{R}^{ 8 \times 36}$.
    We cluster the rows of $\setofinputmat$ and use the result to build $\centroidmat$.
    In this example, 
    $\setofinputmat(0,:)$ and $\setofinputmat(2,:)$ are averaged to $\centroidmat(0,:)$, $\setofinputmat(3,:)$ and $\setofinputmat(6,:)$ are averaged to $\centroidmat(1,:)$, $\setofinputmat(4,:)$ and $\setofinputmat(7,:)$ are averaged to $\centroidmat(2,:)$, and $\setofinputmat(1,:)$ and $\setofinputmat(5,:)$ are averaged to $\centroidmat(3,:)$.
    Lastly, $\centroidmat$ of size $4 \times 36$ is reshaped into 
    $\centroid$ of size $4 \times 3 \times 3 \times 4$.
\end{example}


\noindent 

The clusters in $\seriessmall$ (i.e., reduced locations) may correspond to different numbers of locations in $\series$.
To account for this imbalance, we assign weights $\{w\}_{i=0}^{N_\seriessyn-1}$, where each weight is defined as the ratio between the number of locations contained in the corresponding cluster and the total number of locations in $\series$ (line~\ref{algo:clustering:weight}). 
These weights are used for loss computations in the subsequent distillation process (lines \ref{algo:distill:loss_real} and \ref{algo:distill:loss_syn} of Algorithm~\ref{algo:distill}, and line \ref{algo:train:loss} of Algorithm~\ref{algo:train}).



As shown in the complexity analysis in Section~\ref{sec:method:complexity}, distillation cost grows quadratically with the number of locations.
By replacing the original dataset with its clustered counterpart, this number is reduced, thereby improving both the speed and efficiency of distillation.
Beyond its effectiveness, subset-based granular distillation also reduces the time and space complexity of \method by reducing the number of locations processed at once (see Theorem~\ref{thm:distill} in Section~\ref{sec:method:complexity} for details).

\smallsection{Alternatives of K-means clustering}
K-means can be replaced with alternative clustering methods, such as K-medoids~\cite{zorrilla2024data}.\footnote{Zorrilla et al.~\cite{zorrilla2024data} use K-medoids to cluster locations and train a representative model (spec., ARIMA) for each cluster. During the online phase, a suitable representative model is selected for each query location to make predictions. This strategy avoids training a separate model for every location, thereby reducing overall training cost. Their strategy of training multiple cluster-specific models for selection stands in contrast to our approach that trains a single model (specifically, an STGNN) that jointly incorporates all locations and their spatial relationships using a reduced synthetic dataset.}
Since simple clustering methods may blur local topological structures by ignoring geographical relationships, when geographic information (e.g., longitude and latitude) of locations is given, spatially informed clustering methods~\cite{guo2008regionalization, metis} can be used instead, as explored in Section~\ref{sec:exp:ablation}.

\begin{figure*}[t]
    \centering
    \vspace{-3mm}
    \includegraphics[width=0.90\linewidth]{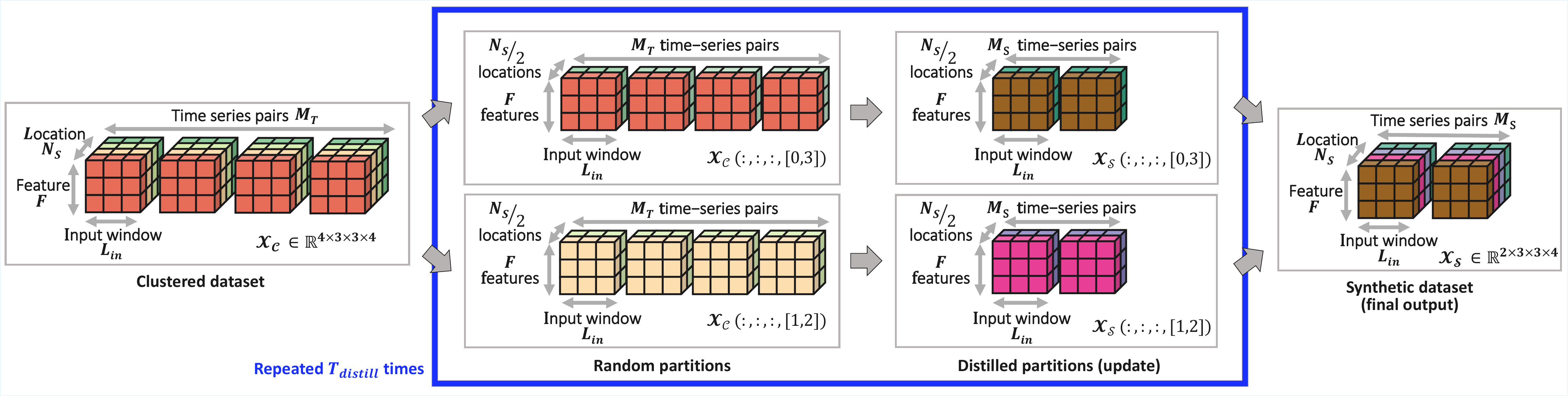}
     \vspace{-2mm}
    \caption{An illustrative example showing how the clustered spatio-temporal dataset (which is averaged within each cluster) are partitioned into two location subsets and distilled at the subset level.
    Entries with the same color correspond to the results for the same location. Refer to Example~\ref{subset:example} for details.}
    \label{fig:augment}
\end{figure*}

\subsection{Subset-based Granular Distillation}\label{sec:method:augmentation}

In this subsection, we present subset-based granular distillation (\textbf{S3} in Section~\ref{sec:method:overview}).
Our approach repeats distillation on fewer, randomly chosen subsets of locations. This enables weaker but important inter-location correlations, which are often overlooked by simultaneous distillation over all locations (\textbf{C3} in Section~\ref{sec:method:overview}), to be effectively reflected in the synthetic dataset.
Thus, the quality of the synthetic dataset is improved, as we empirically show in Section \ref{sec:exp:ablation}.
Below, we describe the details of subset-based granular distillation.

At each distillation iteration, we randomly partition the set of locations $\{0,...,N_\seriessyn-1\}$ of the clustered dataset $\seriessmall$ (or the synthetic dataset $\seriessyn$) into $K$ non-overlapping subsets $\{D_k\}_{k=0}^{K-1}$ of equal size $\frac{N_\seriessyn}{K}$ (line \ref{algo:distill:partition} in Algorithm \ref{algo:distill}; see also Example~\ref{subset:example}).
We re-partition the locations at every iteration to form a new set of different subsets to capture diverse spatial correlations among locations throughout the distillation process.
\begin{example}
\label{subset:example}
  In Figure~\ref{fig:augment}, we provide a partitioning example, where the locations are divided into two subsets ($K=2$):
  $D_0=\{0,3\}$ and $D_1=\{1,2\}$. 
  $\centroid$ and $\centroidfuture$ are split accordingly, and each subset pair 
  $(\centroid, \centroidfuture)(:,:,:,D_k)$ is used to distill the corresponding 
  synthetic subset $(\setofinputsyn, \setoffuturesyn)(:,:,:,D_k)$.
\end{example}

The partitioned subsets are used in four ways.
First, they are used to construct the subset-level inputs from the original inputs (i.e. $\mathcal{X}^B_C$, $\setofinputsyn$, $I_C$, and $I_S$) for the surrogate model.
Second, the surrogate model $f_\theta$, which receives these subset-level inputs, produces subset-level outputs, $\Tilde{\mathcal{Y}}_{\seriessmall}$ and $\Tilde{\mathcal{Y}}_{\seriessyn}$ (lines \ref{algo:distill:output_real} and \ref{algo:distill:output_syn}).
Third, they are used to compute the subset-level losses.
For the $k$-th subset $D_k$, the losses are computed within $D_k$ by summing the reconstruction errors of its constituent locations $i \in D_k$ (lines~\ref{algo:distill:loss_real} and \ref{algo:distill:loss_syn}).
Lastly, the gradients of the synthetic and clustered original datasets, computed from these losses, are applied to update the subset-specific synthetic time series $\setofinputsyn(:,:,:,D_k)$ and $\setoffuturesyn(:,:,:,D_k)$ (lines \ref{algo:distill:grad_match} and \ref{algo:distill:backprop}).

\subsection{Complexity Analysis} \label{sec:method:complexity}

In this subsection, we analyze the time and space complexities of the aforedescribed algorithms: (a) model training on synthetic datasets (Algorithm~\ref{algo:train}) and
(b) dataset distillation (Algorithm~\ref{algo:distill}).
The complexity of location clustering is analyzed in Appendix~B~\cite{appendix}.
Key notations commonly used in the analysis are reviewed as follows.
We use $N_{\series}$ and $N_{\seriessyn}$ to denote the numbers of locations in the original and synthetic datasets, respectively.
The numbers of time series in the original and synthetic datasets are denoted by $M_{\series}$ and $M_{\seriessyn}$, respectively.
The input window size is denoted by $L_{in}$, the target window size by $L_{out}$, the number of input features by $F$, and the number of location subsets by $K$.
\textbf{We assume that MTGNN~\cite{mtgnn} equipped with our location encoder is used for both distillation (as the surrogate model $f_\theta$) and training; and
for simplicity, we omit the hyperparameters that are treated as constants.}



\smallsection{Model training on synthetic datasets}\label{sec:method:complexity:train}
We first analyze the complexity of model training using synthetic datasets generated by \method.
Detailed proofs and analyses are provided in Appendix B~\cite{appendix}.

Theorems~\ref{thm:train:time} and \ref{thm:train:space} present the time and space complexity of model training on the synthetic dataset, where $T_{\text{model}}$ denotes the number of training epochs in Algorithm~\ref{algo:train}.

\begin{thm}[Time Complexity of Model Training] \label{thm:train:time}
    Algorithm \ref{algo:train} takes $O(T_{model} M_\seriessyn(\frac{N_\seriessyn^2}{K} + N_\seriessyn L_{in}))$ time.
\end{thm}

\begin{thm}[Space Complexity of Model Training] \label{thm:train:space}
    Algorithm \ref{algo:train} requires $O(N_{\seriessyn} F(L_{in} + L_{out})M_\seriessyn + \frac{N_\seriessyn^2}{K^2})$ space.
\end{thm}

It is noteworthy that both the time and space complexities grow quadratically with $N_\seriessyn$, the number of locations in the synthetic dataset, which demonstrates the importance of our key idea to reduce $N_\seriessyn$ (\textbf{S1} in Section~\ref{sec:method:overview}).

\smallsection{Dataset distillation by \method} \label{sec:method:complexity:distill} 
We analyze the overall data-distillation cost of \method in Theorem~\ref{thm:distill}.
Recall that, in Algorithm~\ref{algo:distill}, $T_{outer}$ denotes the number of outer iterations, $T_{distill}$ denotes the number of iterations for distillation, and $T_{model}$ denotes the number of epochs for training the surrogate model.
Batch size for the clustered dataset is denoted by $B$.
Detailed proofs and analyses are provided in Appendix B~\cite{appendix}.

\begin{thm}[Time Complexity of \method] \label{thm:distill}
    The time complexity of Algorithm~\ref{algo:distill}, the whole distillation process, is     
    $O(N_{\series}N_{\seriessyn}
    (M_{\series}L_{in}F)T_{kmeans}+T_{outer}T_{distill}((T_{model}M_{\seriessyn}+B)(\frac{N_\seriessyn^2}{K}) + (L_{in}(T_{model}M_{\seriessyn}+B) + (M_\seriessyn + B)L_{out}F_{out}) N_\seriessyn))$.
\end{thm}

As for space complexity, the location clustering step requires $O((N_{\series} + N_{\seriessyn}) M_{\series} L_{in} F)$ space, and the remaining distillation steps require $O(\frac{N_\seriessyn^2}{K^2}+N_{\seriessyn}F(L_{in}+L_{out})(M_\seriessyn+M_\series)$).


Both the time and space complexities of \method show the importance of spatial dimension reduction (\textbf{S1} in Section~\ref{sec:method:overview}), which decreases $N_{\seriessyn}$, and subset-based distillation (\textbf{S3}), which increases $K$. Our location clustering strategy (\textbf{S2}) allows the complexities to grow only linearly with $N_{\series}$. 

\section{Experiments}\label{experiments}

\begin{table}[t]
\caption{Dataset statistics.} \label{tab:dataset}
\centering
\begin{tabular}{l | c | c | c | c}
    \toprule
    & \multicolumn{3}{c|}{\textbf{Data Dimensions}}
    & \textbf{\# Total} \\
    & $M$ & $N$ & $F$ & \textbf{Data Points} \\
    \midrule
    \multicolumn{5}{c}{\textbf{Real-world Datasets}} \\
    \midrule
    GBA \cite{fang2025efficient} & $1,977$ & $2,352$ & $1$ & 4,649,904 \\
    GLA \cite{fang2025efficient} & $1,977$ & $3,834$ & $1$ & 7,579,818 \\
    ERA5 \cite{era5} & $2,137$ & $6,561$ & $6$ & 84,125,142 \\
    CAMS \cite{inness2019cams} & $2,556$ & $7,070$ & $6$ & 108,425,520 \\
    CA \cite{fang2025efficient} & $1,977$ & $8,600$ & $1$ & 17,002,200 \\
    \midrule
     \multicolumn{5}{c}{\textbf{Generated Artificial Datasets}} \\
     \midrule
         Minimum & $262,144$ & $2,352$ & $1$ & 616,562,688 \\
    Maximum & $8,388,608$ & $262,144$ & $8,192$ & 5,050,881,540,096 \\
    \bottomrule
    \end{tabular}
\end{table}

    


We perform experiments to answer the following questions:
\begin{enumerate}[leftmargin=6mm]
    \item [Q1.] \textbf{Distillation Performance:} How effective and efficient is training with the synthetic dataset from \method, compared to competing methods?

    \item[Q2.] \textbf{Cross-model Performance:} How effective are synthetic datasets from \method for training various models?

    \item[Q3.] \textbf{Scalability and Speed:} How does \method scale w.r.t. the number of time series, locations, and features in the dataset? How fast is it, compared to competing methods?

    \item[Q4.] \textbf{Ablation Study:} How does each component of \method affect the distillation time and performance?

    \item[Q5.] \textbf{Hyperparameter Analysis:} How does \method perform under varying hyperparameter settings?
\end{enumerate}

\subsection{Experimental Setup}
\label{sec:exp:settings}

\smallsection{Machines}
All experiments were conducted on a workstation with four RTX A6000 GPUs and 512GB of RAM.

\smallsection{Datasets}\label{exp:datasets}
We use five real-world datasets, and their statistics are summarized in Table \ref{tab:dataset}.
The GBA, GLA, and CA datasets are traffic measurements in several locations. 
We use their versions preprocessed and released in \cite{fang2025efficient}, where traffic volume is used as both the input and target variables.
In addition, we use the ERA5 and CAMS datasets provided in \cite{earchdata}.
ERA5~\cite{era5} provides global weather reanalysis data from ECMWF, while CAMS~\cite{inness2019cams} offers atmospheric analysis and forecasts from the Copernicus Atmospheric Monitoring Service.
For both datasets, the 
six most popular variables are used as inputs; the targets are 10U (10 m eastward wind) for ERA5 and CO (carbon monoxide concentration) for CAMS.
We also generate artificial datasets of varying sizes for scalability evaluation.
We split each dataset into a training set (60\%), a validation set (20\%), and a test set (20\%) in chronological order.
The length of the input and target sequences is fixed to 12.

\smallsection{Evaluation metrics}
To evaluate the performance of models trained on synthetic datasets, we use Relative Root Mean Squared Error (Relative RMSE) and Relative Mean Absolute Error (Relative MAE).
They are defined as follows:
\begin{equation}
    \begin{aligned}
        \text{\(\substack{\text{Relative}\\\text{RMSE}}\)} = \sqrt{\frac{\sum_{i=1}^{N} (y_i - \hat{y}_i)^2}{\sum_{i=1}^{N} (y_i - \bar{y})^2}}, 
        \text{\(\substack{\text{Relative}\\\text{MAE}}\)} = \frac{\sum_{i=1}^{N} \lvert y_i - \hat{y}_i \rvert}{\sum_{i=1}^{N} \lvert y_i - \bar{y} \rvert},
    \end{aligned}
\end{equation}
where $y_i$ is the ground truth value, $\hat{y}_i$ is the predicted value, and $\bar{y}$ is the mean of the ground truth values.
Relative metrics are used to normalize differences in scale across datasets, and for these metrics,
smaller values indicate better accuracy.



\smallsection{Baselines}
We employ nine baselines, categorized as follows:
\begin{itemize}[leftmargin=*]
\item \textbf{Coreset selection methods:}
\textbf{Random sampling} randomly draws samples (i.e., input-target time series pairs) from the original spatio-temporal time series dataset $\mathcal{T}$.
\textbf{K-Center}~\cite{farahani2009facility} selects representative samples from each cluster from K-means clustering.
\textbf{Herding}~\cite{welling2009herding} greedily selects samples to best approximate the mean of the original dataset.
\textbf{CRAIG} \cite{mirzasoleiman2020coresets} selects a subset of training data with weights to closely estimate the full gradient.
\item \textbf{Distillation methods for general datasets:} 
\textbf{DC}
~\cite{zhao2021dataset}, \textbf{DM}
~\cite{zhao2023dataset}, \textbf{MTT}
~\cite{cazenavette2022dataset}, 
\textbf{DATM} \cite{guo2023towards},
\textbf{IDM} \cite{zhao2023improved},
and \textbf{Frepo}
~\cite{zhou2022dataset} employ gradient matching, distribution matching, trajectory matching, and meta-learning, respectively, which are detailed in Section~\ref{related works}. 
\item \textbf{Distillation methods for time series datasets:} \textbf{CondTSF}
~\cite{ding2024condtsf} and \textbf{TimeDC}
~\cite{miao2024less}, which are described in Section \ref{related works}, are specialized dataset-distillation methods for univariate and multivariate time series datasets, respectively.
\end{itemize}

\smallsection{Distillation details}
For all baselines, we use or adapt the implementations provided by the authors.
For \method and its variants, we use MTGNN~\cite{mtgnn} augmented with our proposed location encoder as the surrogate model $f_\theta$.
For DC, DM, MTT, and Frepo, we also use MTGNN but without the location encoder, as it is unnecessary when the number of locations is not reduced.
For CondTSF and TimeDC, we adopt the same surrogate models as specified in their papers.
We use the target feature as the sole input feature in TimeDC and CondTSF, as they support only a single input variable.

For all methods, we set the number of outer iterations ($T_{outer}$) to 200 for the GBA/GLA datasets and 100 for the CA/ERA5/CAMS datasets.
We use 20 distillation iterations ($T_{distill}$), 10 model training epochs ($T_{model}$), and maximum of 300 iterations ($T_{kmeans}$) for K-means clustering in \method.
For distillation methods which reduce the number of locations, when the compression ratio is 0.5\%, we set the number of time series ($M_\seriessyn$) and locations ($N_\seriessyn$) to 10\% and 5\% of the original dataset, respectively, and to 10\% and 10\% when the compression ratio is 1\% (see Section~\ref{sec:exp:hyperparam} for a related anaylsis).
For the baselines in Section \ref{sec:exp:main}, to match the compression ratios, we compress only the number of time series ($M_\seriessyn$) as they do not inherently support compression of the location count ($N_\seriessyn$).
We set the number of location subsets ($K$) to $4$ and the embedding dimension for the location encoder model to $32$.
In Sections \ref{sec:exp:main} and \ref{sec:exp:cross} each dataset is distilled five times with different seeds per method, and each model is trained with five random initial parameters per distilled data, yielding 25 experiments in total.
The learning rates used to train the surrogate model ($\eta$) and to update the synthetic data are tuned via grid search over \(\{10^{-2}, 10^{-3}, 10^{-4}\}\).
Refer to \cite{appendix} for more details and hyperparameter settings of the baselines.

\smallsection{Inference details}
For inference after training, each batch consists of 8 time series. When the number of locations exceeds $2^{15}$, we further split the data into batches of $2^{15}$ locations to avoid out-of-memory errors in STGNNs.




\begin{table*}[!t]
  \vspace{-2mm}
  \centering
  \caption{Effectiveness of dataset-distillation methods. The synthetic dataset from each method is used to train MTGNN, equipped with the location encoders when necessary, for evaluation. The performances of our proposed method are highlighted in bold; the best performances in each setting are highlighted in yellow. 
  O.O.M.: out of memory. \label{tab:results}} 
  
  \subfloat[Compression ratio: 0.5\%]{
      \resizebox{\linewidth}{!}{
      \begin{tabular}{l||cc||cc||cc||cc||cc}
        \toprule 
        & \multicolumn{2}{c||}{\textbf{GBA}} 
        & \multicolumn{2}{c||}{\textbf{GLA}} 
        & \multicolumn{2}{c||}{\textbf{ERA5}} 
        & \multicolumn{2}{c||}{\textbf{CA}} 
        & \multicolumn{2}{c}{\textbf{CAMS}} \\
        
        \midrule
        \multirow{2}{*}{\textbf{Method}} & Relative & Relative & Relative & Relative & Relative & Relative & Relative & Relative & Relative & Relative  \\
        & MAE & RMSE & MAE & RMSE &  MAE & RMSE & MAE &  RMSE & MAE & RMSE \\
        \midrule    
         Random & 0.380 $\pm$ 0.095 & 0.453 $\pm$ 0.086 & 0.313 $\pm$ 0.050 & 0.410 $\pm$ 0.051 & 0.482 $\pm$ 0.036 & 0.520 $\pm$ 0.028 & 0.323 $\pm$ 0.030 & 0.431 $\pm$ 0.065 & 0.660 $\pm$ 0.104 & 0.796 $\pm$ 0.109 \\
         K-Center~\cite{farahani2009facility} & 0.319 $\pm$ 0.001 & 0.400 $\pm$ 0.042 & 0.269 $\pm$ 0.002 & 0.356 $\pm$ 0.028 & 0.461 $\pm$ 0.011 & 0.524 $\pm$ 0.012 & 0.289 $\pm$ 0.051 & 0.355 $\pm$ 0.034 & 0.883 $\pm$ 0.836 & 0.883 $\pm$ 0.073 \\
         Herding~\cite{welling2009herding} & 0.467 $\pm$ 0.008 & 0.506 $\pm$ 0.004 & 0.331 $\pm$ 0.013 & 0.428 $\pm$ 0.004 & 0.471 $\pm$ 0.000 & 0.522 $\pm$ 0.001 & 0.472 $\pm$ 0.000 & 0.521 $\pm$ 0.004 & 0.584 $\pm$ 0.000 & 0.730 $\pm$ 0.002 \\
        CRAIG \cite{mirzasoleiman2020coresets} & 0.291 $\pm$ 0.007 & 0.354 $\pm$ 0.007 & 0.258 $\pm$ 0.015 & 0.323 $\pm$ 0.017 & 0.430 $\pm$ 0.004 & 0.470 $\pm$ 0.003 & 0.287 $\pm$ 0.020 & 0.354 $\pm$ 0.021 & 0.563 $\pm$ 0.006 & 0.685 $\pm$ 0.008 \\
        
        \midrule
        DC~\cite{zhao2021dataset} & 0.281 $\pm$ 0.009 & 0.348 $\pm$ 0.006 & 0.243 $\pm$ 0.009 & 0.311 $\pm$ 0.009 & 0.392 $\pm$ 0.004 & 0.442 $\pm$ 0.004 & 0.254 $\pm$ 0.011 & 0.331 $\pm$ 0.014 & 0.630 $\pm$ 0.005 & 0.735 $\pm$ 0.006 \\
        DM~\cite{zhao2023dataset} & 0.338 $\pm$ 0.021 & 0.409 $\pm$ 0.051 & 0.298 $\pm$ 0.023 & 0.380 $\pm$ 0.025 & 0.420 $\pm$ 0.007 & 0.467 $\pm$ 0.008 & 0.321 $\pm$ 0.027 & 0.405 $\pm$ 0.028 & 0.528 $\pm$ 0.020 & 0.674 $\pm$ 0.011 \\
        MTT~\cite{cazenavette2022dataset} & 0.339 $\pm$ 0.051 & 0.414 $\pm$ 0.055 & 0.343 $\pm$ 0.058 & 0.439 $\pm$ 0.066 & O.O.M & O.O.M & O.O.M & O.O.M & O.O.M & O.O.M \\
        DATM~\cite{guo2023towards} & 0.430 $\pm$ 0.012 & 0.517 $\pm$ 0.007 & 0.332 $\pm$ 0.016 & 0.402 $\pm$ 0.022 & O.O.M & O.O.M & O.O.M & O.O.M & O.O.M & O.O.M \\
        IDM~\cite{zhao2023improved} & 0.358 $\pm$ 0.062 & 0.420 $\pm$ 0.050 & 0.300 $\pm$ 0.017 & 0.381 $\pm$ 0.024 & 0.438 $\pm$ 0.018 & 0.481 $\pm$ 0.018 & 0.315 $\pm$ 0.037 & 0.394 $\pm$ 0.047 & 0.567 $\pm$ 0.054 & 0.709 $\pm$ 0.049 \\
        Frepo~\cite{zhou2022dataset} & 0.397 $\pm$ 0.008 & 0.465 $\pm$ 0.075 & 0.350 $\pm$ 0.087 & 0.439 $\pm$ 0.084 & 0.482 $\pm$ 0.036 & 0.510 $\pm$ 0.031 & 0.361 $\pm$ 0.085 & 0.436 $\pm$ 0.084 & 0.688 $\pm$ 0.116 & 0.802 $\pm$ 0.118\\
    
        \midrule
        CondTSF~\cite{ding2024condtsf} & 0.337 $\pm$ 0.040 & 0.407 $\pm$ 0.041 & 0.311 $\pm$ 0.027 & 0.394 $\pm$ 0.031 & 0.449 $\pm$ 0.017 & 0.489 $\pm$ 0.017 & 0.322 $\pm$ 0.050 & 0.403 $\pm$ 0.051 & 0.555 $\pm$ 0.053 & 0.692 $\pm$ 0.037 \\
        TimeDC~\cite{miao2024less} & 0.350 $\pm$ 0.031 & 0.422 $\pm$ 0.030 & 0.344 $\pm$ 0.061 & 0.445 $\pm$ 0.068 & 0.437 $\pm$ 0.015 & 0.480 $\pm$ 0.015 & 0.360 $\pm$ 0.073 & 0.448 $\pm$ 0.074 & 0.585 $\pm$ 0.105 & 0.729 $\pm$ 0.078 \\

        \midrule 
        \method-K-Center & 0.305 $\pm$ 0.014 & 0.367 $\pm$ 0.006 & 0.248 $\pm$ 0.007 & 0.313 $\pm$ 0.008 & 0.396 $\pm$ 0.003 & 0.440 $\pm$ 0.005 & 0.291 $\pm$ 0.012 & 0.348 $\pm$ 0.009 & 0.903 $\pm$ 0.093 & 0.884 $\pm$ 0.077  \\
        \method-DC & 0.263 $\pm$ 0.007 & 0.325 $\pm$ 0.006 & 0.227 $\pm$ 0.002 & 0.292 $\pm$ 0.006 & 0.392 $\pm$ 0.002 & 0.431 $\pm$ 0.002 & 0.247 $\pm$ 0.006 & 0.314 $\pm$ 0.005 & 0.545 $\pm$ 0.007 & 0.668 $\pm$ 0.008 \\
        \method-MTT & 0.263 $\pm$ 0.007 & 0.325 $\pm$ 0.004 & 0.227 $\pm$ 0.004 & 0.300 $\pm$ 0.004 & 0.399 $\pm$ 0.009 & 0.437 $\pm$ 0.007 & 0.250 $\pm$ 0.004 & 0.311 $\pm$ 0.006 & 0.570 $\pm$ 0.010 & 0.685 $\pm$ 0.010 \\
        \method-CondTSF & 0.260 $\pm$ 0.006 & 0.335 $\pm$ 0.010 & 0.225 $\pm$ 0.010 & 0.298 $\pm$ 0.005 & 0.407 $\pm$ 0.002 & 0.452 $\pm$ 0.002 & 0.247 $\pm$ 0.008 & 0.319 $\pm$ 0.013 & \cellcolor{yellow} 0.514 $\pm$ 0.003 & \cellcolor{yellow} 0.648 $\pm$ 0.002 \\
        \midrule
        \textbf{\makecell[l]{
        \method \\
        \method\ (1 feat.)}
        }
        & \makecell{\colorbox{yellow}{\centering\textbf{0.251 $\pm$ 0.005}}} 
        & \makecell{\colorbox{yellow}{\centering\textbf{0.317 $\pm$ 0.005}}} 
        & \makecell{\colorbox{yellow}{\centering\textbf{0.211 $\pm$ 0.008}}} 
        & \makecell{\colorbox{yellow}{\centering\textbf{0.277 $\pm$ 0.002}}} 
        & \makecell{\colorbox{yellow}{{\centering\textbf{0.385 $\pm$ 0.007}}} \\ \footnotesize \textbf{0.400 $\pm$ 0.003}}
        & \makecell{\colorbox{yellow}{{\centering\textbf{0.426 $\pm$ 0.004}}} \\ \footnotesize \textbf{0.447 $\pm$ 0.003}}
        & \makecell{\colorbox{yellow}{\centering\textbf{0.226 $\pm$ 0.001}}} 
        & \makecell{\colorbox{yellow}{\centering\textbf{0.292 $\pm$ 0.002}}} 
        & \makecell{{{\centering\textbf{0.522 $\pm$ 0.005}}} \\ \footnotesize \textbf{0.520 $\pm$ 0.007}}
        & \makecell{{{\centering\textbf{0.653 $\pm$ 0.005}}} \\ \footnotesize \textbf{0.649 $\pm$ 0.007}}
        \\        
        \midrule
        Original data
        & 0.207 $\pm$ 0.008 & 0.264 $\pm$ 0.010 & 0.182 $\pm$ 0.001 & 0.250 $\pm$ 0.005 & 0.368 $\pm$ 0.004 & 0.424 $\pm$ 0.010 & 0.120 $\pm$ 0.002 & 0.253 $\pm$ 0.003 & 0.500 $\pm$ 0.005 & 0.661 $\pm$ 0.004 \\
        \bottomrule
      \end{tabular}
      }
    }

    \par\vspace{-2mm}

    \subfloat[Compression ratio: 1\%]{
      \resizebox{\linewidth}{!}{
      \begin{tabular}{l||cc||cc||cc||cc||cc}
        \toprule 
        & \multicolumn{2}{c||}{\textbf{GBA}} 
        & \multicolumn{2}{c||}{\textbf{GLA}} 
        & \multicolumn{2}{c||}{\textbf{ERA5}} 
        & \multicolumn{2}{c||}{\textbf{CA}} 
        & \multicolumn{2}{c}{\textbf{CAMS}} \\
        
        \midrule
        \multirow{2}{*}{\textbf{Method}} & Relative & Relative & Relative & Relative & Relative & Relative & Relative & Relative & Relative & Relative  \\
        & MAE & RMSE & MAE & RMSE &  MAE & RMSE & MAE &  RMSE & MAE & RMSE \\
        \midrule    
         Random & 0.291 $\pm$ 0.016 & 0.364 $\pm$ 0.017 & 0.271 $\pm$ 0.023 & 0.355 $\pm$ 0.033 & 0.453 $\pm$ 0.012 & 0.494 $\pm$ 0.009 & 0.289 $\pm$ 0.031 & 0.374 $\pm$ 0.045 & 0.617 $\pm$ 0.039 & 0.737 $\pm$ 0.041 \\
         K-Center~\cite{farahani2009facility} & 0.294 $\pm$ 0.001 & 0.366 $\pm$ 0.003  & 0.240 $\pm$ 0.001 & 0.309 $\pm$ 0.002 & 0.434 $\pm$ 0.000 & 0.480 $\pm$ 0.000 & 0.256 $\pm$ 0.001 & 0.326 $\pm$ 0.002 & 0.682 $\pm$ 0.010 & 0.841 $\pm$ 0.009 \\
         Herding~\cite{welling2009herding} & 0.312 $\pm$ 0.001 & 0.387 $\pm$ 0.001 &  0.273 $\pm$ 0.000 & 0.342 $\pm$ 0.001 & 0.441 $\pm$ 0.000 & 0.478 $\pm$ 0.001 & 0.243 $\pm$ 0.001 & 0.313 $\pm$ 0.001 & 0.567 $\pm$ 0.000 & 0.696 $\pm$ 0.000 \\
        CRAIG \cite{mirzasoleiman2020coresets} & 0.272 $\pm$ 0.007 & 0.339 $\pm$ 0.006 & 0.243 $\pm$ 0.005 & 0.311 $\pm$ 0.005 & 0.417 $\pm$ 0.011 & 0.457 $\pm$ 0.009 & 0.261 $\pm$ 0.005 & 0.323 $\pm$ 0.005 & 0.551 $\pm$ 0.010 & 0.679 $\pm$ 0.012 \\
        
        \midrule
        DC~\cite{zhao2021dataset} & 0.272 $\pm$ 0.005 & 0.340 $\pm$ 0.006 & 0.236 $\pm$ 0.003 & 0.304 $\pm$ 0.001 & 0.393 $\pm$ 0.002 & 0.434 $\pm$ 0.002 & 0.253 $\pm$ 0.003 & 0.318 $\pm$ 0.002 & 0.520 $\pm$ 0.009 & {0.652 $\pm$ 0.008} \\
        DM~\cite{zhao2023dataset} & 0.296 $\pm$ 0.013 & 0.361 $\pm$ 0.010 & 0.279 $\pm$ 0.028  & 0.356 $\pm$ 0.032 & 0.417 $\pm$ 0.017 & 0.461 $\pm$ 0.016 & 0.285 $\pm$ 0.022 & 0.361 $\pm$ 0.031 & 0.538 $\pm$ 0.018 & 0.678 $\pm$ 0.021 \\
        MTT~\cite{cazenavette2022dataset}& 0.289 $\pm$ 0.015 & 0.357 $\pm$ 0.013 & O.O.M & O.O.M & O.O.M & O.O.M & O.O.M & O.O.M & O.O.M & O.O.M \\
        DATM~\cite{guo2023towards} & 0.376 $\pm$ 0.014 & 0.463 $\pm$ 0.014 & 0.267 $\pm$ 0.008 & 0.336 $\pm$ 0.003 & O.O.M & O.O.M & O.O.M & O.O.M & O.O.M & O.O.M \\
        IDM~\cite{zhao2023improved} & 0.290 $\pm$ 0.018 & 0.356 $\pm$ 0.018 & 0.280 $\pm$ 0.009 & 0.358 $\pm$ 0.008 & 0.411 $\pm$ 0.013 & 0.457 $\pm$ 0.012 & 0.283 $\pm$ 0.021 & 0.349 $\pm$ 0.019 & 0.527 $\pm$ 0.008 & 0.666 $\pm$ 0.008 \\
        Frepo~\cite{zhou2022dataset} & 0.283 $\pm$ 0.002 & 0.356 $\pm$ 0.006 & 0.256 $\pm$ 0.012 & 0.330 $\pm$ 0.019 & 0.419 $\pm$ 0.012 & 0.460 $\pm$ 0.013 & 0.259 $\pm$ 0.011 & 0.329 $\pm$ 0.013 & 0.653 $\pm$ 0.007 & 0.748 $\pm$ 0.002 \\
    
        \midrule
        CondTSF~\cite{ding2024condtsf} & 0.294 $\pm$ 0.014 & 0.364 $\pm$ 0.012 & 0.274 $\pm$ 0.024 & 0.356 $\pm$ 0.033 & 0.413 $\pm$ 0.005 & 0.461 $\pm$ 0.005 & 0.279 $\pm$ 0.018 & 0.359 $\pm$ 0.029 & 0.534 $\pm$ 0.032 & 0.680 $\pm$ 0.028 \\
        TimeDC~\cite{miao2024less} & 0.295 $\pm$ 0.017 & 0.367 $\pm$ 0.016 & 0.274 $\pm$ 0.028 & 0.358 $\pm$ 0.040 & 0.426 $\pm$ 0.013 & 0.472 $\pm$ 0.014 & 0.295 $\pm$ 0.036 & 0.377 $\pm$ 0.052 & 0.560 $\pm$ 0.054 & 0.689 $\pm$ 0.021 \\

        \midrule
        \method-K-Center & 0.280 $\pm$ 0.003 & 0.340 $\pm$ 0.006 & 0.243 $\pm$ 0.007 & 0.302 $\pm$ 0.009 & 0.395 $\pm$ 0.004 & 0.436 $\pm$ 0.005 & 0.274 $\pm$ 0.008 & 0.332 $\pm$ 0.011 & 0.872 $\pm$ 0.106 & 0.866 $\pm$ 0.066 \\
        \method-DC & 0.258 $\pm$ 0.010 & 0.328 $\pm$ 0.009 & 0.223 $\pm$ 0.004 & 0.295 $\pm$ 0.007 & 0.392 $\pm$ 0.003 & 0.432 $\pm$ 0.003 & 0.240 $\pm$ 0.004 & 0.305 $\pm$ 0.003 & 0.537 $\pm$ 0.008 & 0.663 $\pm$ 0.002 \\
        \method-MTT & 0.261 $\pm$ 0.005 & 0.321 $\pm$ 0.004 & 0.224 $\pm$ 0.007 & 0.293 $\pm$ 0.003 & 0.403 $\pm$ 0.003 & 0.444 $\pm$ 0.004 & 0.248 $\pm$ 0.012 & 0.317 $\pm$ 0.014 & 0.550 $\pm$ 0.016 & 0.681 $\pm$ 0.005 \\
        \method-CondTSF & 0.261 $\pm$ 0.010 & 0.321 $\pm$ 0.003 & 0.220 $\pm$ 0.005 & 0.289 $\pm$ 0.004 & 0.405 $\pm$ 0.002 & 0.449 $\pm$ 0.002 & 0.243 $\pm$ 0.006 & 0.313 $\pm$ 0.004 & 0.513 $\pm$ 0.007 & 0.645 $\pm$ 0.002 \\
        \midrule
        \textbf{\makecell[l]{
        \method \\
        \method\ (1 feat.)
        }}
        & \makecell{\colorbox{yellow}{\centering\textbf{0.242 $\pm$ 0.001}}} 
        & \makecell{\colorbox{yellow}{\centering\textbf{0.307 $\pm$ 0.002}}} 
        & \makecell{\colorbox{yellow}{\centering\textbf{0.205 $\pm$ 0.001}}} 
        & \makecell{\colorbox{yellow}{\centering\textbf{0.272 $\pm$ 0.001}}} 
        & \makecell{\colorbox{yellow}{{\centering\textbf{0.384 $\pm$ 0.005}}} \\ \footnotesize \textbf{0.399 $\pm$ 0.003}}
        & \makecell{\colorbox{yellow}{{\centering\textbf{0.426 $\pm$ 0.004}}} \\ \footnotesize \textbf{0.445 $\pm$ 0.004}}
        & \makecell{\colorbox{yellow}{\centering\textbf{0.222 $\pm$ 0.002}}} 
        & \makecell{\colorbox{yellow}{\centering\textbf{0.287 $\pm$ 0.002}}} 
        & \makecell{{{\centering\textbf{0.518 $\pm$ 0.008}}} \\ \colorbox{yellow}{\footnotesize \textbf{0.501 $\pm$ 0.006}}}
        & \makecell{{{\centering\textbf{0.652 $\pm$ 0.006}}} \\ \colorbox{yellow}{\footnotesize \textbf{0.644 $\pm$ 0.003}}}      
        \\        
        \midrule
        Original data
        & 0.207 $\pm$ 0.008 & 0.264 $\pm$ 0.010 & 0.182 $\pm$ 0.001 & 0.250 $\pm$ 0.005 & 0.368 $\pm$ 0.004 & 0.424 $\pm$ 0.010 & 0.120 $\pm$ 0.002 & 0.253 $\pm$ 0.003 & 0.500 $\pm$ 0.005 & 0.661 $\pm$ 0.004 \\
        
        \bottomrule
      \end{tabular}}
    }
\end{table*}
\begin{figure*}[t]
    \vspace{-4mm}
    \centering
        \includegraphics[width=0.35\linewidth]{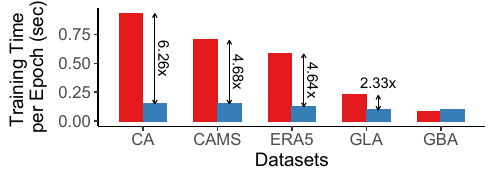}
        \includegraphics[width=0.35\linewidth]{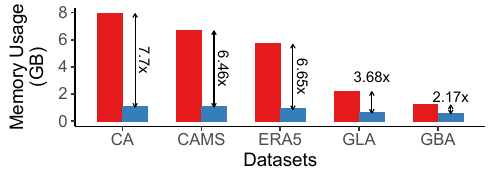}
        \includegraphics[width=0.085\linewidth]{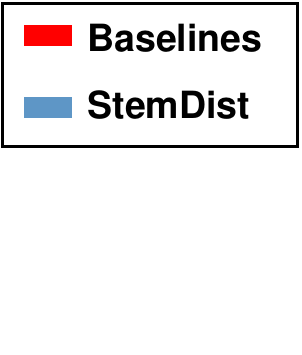}
    \vspace{-3mm}
    \caption{Training on synthetic datasets distilled by \method is faster and more memory-efficient than training on those distilled by the baselines.}
    \label{fig:comparision}
\end{figure*}

\begin{table*}[t]
    \vspace{-2mm}
  \centering
  \caption{Cross-model generalization of dataset-distillation methods. The synthetic dataset produced by each method is used to train various machine learning models for evaluation.
  The performances of our methods are highlighted in bold; and the best performances in each setting are highlighted in yellow.} \label{tab:cross-archi}
  \resizebox{\textwidth}{!}{
  \begin{tabular}{l||cc||cc||cc||cc||cc}
    \toprule 
    \multicolumn{11}{c}{\textbf{Graph wavenet~\cite{wu2019graph}}}\\
   \midrule
    & \multicolumn{2}{c||}{\textbf{GBA}} 
    & \multicolumn{2}{c||}{\textbf{GLA}} 
    & \multicolumn{2}{c||}{\textbf{ERA5}} 
    & \multicolumn{2}{c||}{\textbf{CA}} 
    & \multicolumn{2}{c}{\textbf{CAMS}} \\
    
    \midrule
    \multirow{2}{*}{\textbf{Method}} & Relative & Relative & Relative & Relative & Relative & Relative & Relative & Relative & Relative & Relative  \\
    & MAE & RMSE & MAE & RMSE &  MAE & RMSE & MAE &  RMSE & MAE & RMSE \\
    \midrule         
     K-Center~\cite{farahani2009facility} & 0.336 $\pm$ 0.027 & 0.427 $\pm$ 0.032 & 0.279 $\pm$ 0.021 & 0.367 $\pm$ 0.018 & 0.466 $\pm$ 0.017 & 0.510 $\pm$ 0.012 & 0.318 $\pm$ 0.032 & 0.396 $\pm$ 0.030 & 0.763 $\pm$ 0.055 & 0.796 $\pm$ 0.028 \\
    \midrule
    DC~\cite{zhao2021dataset} & 0.321 $\pm$ 0.036 & 0.395 $\pm$ 0.033 & 0.288 $\pm$ 0.031 & 0.312 $\pm$ 0.009 & 0.424 $\pm$ 0.007 & 0.470 $\pm$ 0.004 & 0.292 $\pm$ 0.043 & 0.369 $\pm$ 0.038 & 0.550 $\pm$ 0.006 & 0.688 $\pm$ 0.004 \\
    CondTSF~\cite{ding2024condtsf} & 0.346 $\pm$ 0.065 & 0.416 $\pm$ 0.066 & 0.302 $\pm$ 0.030 & 0.387 $\pm$ 0.028 & 0.464 $\pm$ 0.008 & 0.514 $\pm$ 0.007 & 0.322 $\pm$ 0.038 & 0.413 $\pm$ 0.031 & 0.531 $\pm$ 0.026 & 0.695 $\pm$ 0.022 \\
    \textbf{\method} 
    & \makecell{\colorbox{yellow}{\centering\textbf{0.241 $\pm$ 0.007}}}
    & \makecell{\colorbox{yellow}{\centering\textbf{0.310 $\pm$ 0.007}}}
    & \makecell{\colorbox{yellow}{\centering\textbf{0.212 $\pm$ 0.006}}}
    & \makecell{\colorbox{yellow}{\centering\textbf{0.273 $\pm$ 0.003}}}
    & \makecell{\colorbox{yellow}{\centering\textbf{0.385 $\pm$ 0.004}}}
    & \makecell{\colorbox{yellow}{\centering\textbf{0.431 $\pm$ 0.003}}}
    & \makecell{\colorbox{yellow}{\centering\textbf{0.222 $\pm$ 0.004}}}
    & \makecell{\colorbox{yellow}{\centering\textbf{0.293 $\pm$ 0.006}}}
    & \makecell{\colorbox{yellow}{\centering\textbf{0.528 $\pm$ 0.006}}}
    & \makecell{\colorbox{yellow}{\centering\textbf{0.665 $\pm$ 0.007}}}
    \\
    \midrule
    Original data & 0.181 $\pm$ 0.005 & 0.227 $\pm$ 0.009 & 0.166 $\pm$ 0.007 & 0.212 $\pm$ 0.011 & 0.376 $\pm$ 0.007 & 0.419 $\pm$ 0.005 & 0.169 $\pm$ 0.009 & 0.215 $\pm$ 0.007 & 0.488 $\pm$ 0.014 & 0.631 $\pm$ 0.007 \\    
    \bottomrule
    \toprule

    \multicolumn{11}{c}{\textbf{STGCN~\cite{stgcn}}}\\
    \midrule
    & \multicolumn{2}{c||}{\textbf{GBA}} 
    & \multicolumn{2}{c||}{\textbf{GLA}} 
    & \multicolumn{2}{c||}{\textbf{ERA5}} 
    & \multicolumn{2}{c||}{\textbf{CA}} 
    & \multicolumn{2}{c}{\textbf{CAMS}} \\
    \midrule

    \multirow{2}{*}{\textbf{Method}} & Relative & Relative & Relative & Relative & Relative & Relative & Relative & Relative & Relative & Relative  \\
    & MAE & RMSE & MAE & RMSE &  MAE & RMSE & MAE &  RMSE & MAE & RMSE \\
    \midrule         
    K-Center\cite{farahani2009facility} & 0.606 $\pm$ 0.091 & 0.714 $\pm$ 0.085 & 0.643 $\pm$ 0.094 & 0.746 $\pm$ 0.062 & 0.791 $\pm$ 0.081 & 0.800 $\pm$ 0.046 & 0.625 $\pm$ 0.046 & 0.731 $\pm$ 0.073 & 1.155 $\pm$ 0.045 & 1.075 $\pm$ 0.019 \\
    \midrule
    DC~\cite{zhao2021dataset} & 0.767 $\pm$ 0.077 & 0.812 $\pm$ 0.053 & 0.767 $\pm$ 0.078 & 0.866 $\pm$ 0.031 & 0.733 $\pm$ 0.044 & 0.753 $\pm$ 0.060 & 0.655 $\pm$ 0.088 & 0.736 $\pm$ 0.068 & 0.936 $\pm$ 0.035 & 0.986 $\pm$ 0.018 \\
    CondTSF~\cite{ding2024condtsf} & 0.628 $\pm$ 0.065 & 0.697 $\pm$ 0.038 & 0.622 $\pm$ 0.070 & 0.734 $\pm$ 0.087 & 0.762 $\pm$ 0.065 & 0.796 $\pm$ 0.052 & 0.616 $\pm$ 0.095 & 0.730 $\pm$ 0.094 & 0.896 $\pm$ 0.069 & 0.950 $\pm$ 0.037 \\
    \textbf{\method}
    & \makecell{\colorbox{yellow}{\centering\textbf{0.441 $\pm$ 0.038}}}
    & \makecell{\colorbox{yellow}{\centering\textbf{0.501 $\pm$ 0.041}}}
    & \makecell{\colorbox{yellow}{\centering\textbf{0.392 $\pm$ 0.032}}}
    & \makecell{\colorbox{yellow}{\centering\textbf{0.459 $\pm$ 0.034}}}
    & \makecell{\colorbox{yellow}{\centering\textbf{0.632 $\pm$ 0.032}}}
    & \makecell{\colorbox{yellow}{\centering\textbf{0.667 $\pm$ 0.037}}}
    & \makecell{\colorbox{yellow}{\centering\textbf{0.443 $\pm$ 0.050}}}
    & \makecell{\colorbox{yellow}{\centering\textbf{0.504 $\pm$ 0.045}}}
    & \makecell{\colorbox{yellow}{\centering\textbf{0.748 $\pm$ 0.022}}}
    & \makecell{\colorbox{yellow}{\centering\textbf{0.844 $\pm$ 0.011}}}
    \\
    \midrule
    Original data & 0.247 $\pm$ 0.008 & 0.294 $\pm$ 0.008 & 0.240 $\pm$ 0.005 &  0.295 $\pm$ 0.003 & 0.511 $\pm$ 0.006 & 0.546 $\pm$ 0.012 & 0.371 $\pm$ 0.348 & 0.268 $\pm$ 0.006 & 0.667 $\pm$ 0.020 & 0.743 $\pm$ 0.010\\    
    \bottomrule
    \toprule

    \multicolumn{11}{c}{\textbf{FourierGNN~\cite{fouriergnn}}}\\
    \midrule
    & \multicolumn{2}{c||}{\textbf{GBA}} 
    & \multicolumn{2}{c||}{\textbf{GLA}} 
    & \multicolumn{2}{c||}{\textbf{ERA5}} 
    & \multicolumn{2}{c||}{\textbf{CA}} 
    & \multicolumn{2}{c}{\textbf{CAMS}} \\
    \midrule

    \multirow{2}{*}{\textbf{Method}} & Relative & Relative & Relative & Relative & Relative & Relative & Relative & Relative & Relative & Relative  \\
    & MAE & RMSE & MAE & RMSE &  MAE & RMSE & MAE &  RMSE & MAE & RMSE \\
    \midrule         
    K-Center\cite{farahani2009facility} & 0.353 $\pm$ 0.032 & 0.419 $\pm$ 0.027 & 0.310 $\pm$ 0.027 & 0.395 $\pm$ 0.025 & 0.432 $\pm$ 0.005 & 0.480 $\pm$ 0.005 & 0.360 $\pm$ 0.020 & 0.437 $\pm$ 0.021 & 0.651 $\pm$ 0.012 & 0.795 $\pm$ 0.021 \\
    \midrule
    DC~\cite{zhao2021dataset} & 0.335 $\pm$ 0.015 & 0.403 $\pm$ 0.015 & 0.301 $\pm$ 0.020 & 0.376 $\pm$ 0.021 & 0.426 $\pm$ 0.005 & 0.469 $\pm$ 0.003 & 0.337 $\pm$ 0.010 & 0.409 $\pm$ 0.011 & 0.561 $\pm$ 0.024 & 0.703 $\pm$ 0.028 \\
    CondTSF~\cite{ding2024condtsf} & 0.348 $\pm$ 0.021 & 0.419 $\pm$ 0.036 & 0.313 $\pm$ 0.018 & 0.421 $\pm$ 0.009 & 0.469 $\pm$ 0.022 & 0.511 $\pm$ 0.019 & 0.351 $\pm$ 0.028 & 0.441 $\pm$ 0.023 & 0.557 $\pm$ 0.059 & 0.715 $\pm$ 0.070 \\
    \textbf{\method}
    & \makecell{\colorbox{yellow}{\centering\textbf{0.309 $\pm$ 0.003}}}
    & \makecell{\colorbox{yellow}{\centering\textbf{0.384 $\pm$ 0.002}}}
    & \makecell{\colorbox{yellow}{\centering\textbf{0.263 $\pm$ 0.003}}}
    & \makecell{\colorbox{yellow}{\centering\textbf{0.339 $\pm$ 0.004}}}
    & \makecell{\colorbox{yellow}{\centering\textbf{0.412 $\pm$ 0.001}}}
    & \makecell{\colorbox{yellow}{\centering\textbf{0.459 $\pm$ 0.001}}}
    & \makecell{\colorbox{yellow}{\centering\textbf{0.293 $\pm$ 0.007}}}
    & \makecell{\colorbox{yellow}{\centering\textbf{0.371 $\pm$ 0.006}}}
    & \makecell{\colorbox{yellow}{\centering\textbf{0.534 $\pm$ 0.008}}}
    & \makecell{\colorbox{yellow}{\centering\textbf{0.658 $\pm$ 0.001}}}
    \\
    \midrule
    Original data & 0.248 $\pm$ 0.006 & 0.307 $\pm$ 0.012 & 0.216 $\pm$ 0.009 & 0.284 $\pm$ 0.006 & 0.396 $\pm$ 0.002 & 0.444 $\pm$ 0.002 & 0.219 $\pm$ 0.009 & 0.288 $\pm$ 0.006 & 0.496 $\pm$ 0.003 & 0.642 $\pm$ 0.002 \\   
    \bottomrule
    
  \end{tabular}
  }
\end{table*}

\subsection{Distillation Performance} \label{sec:exp:main}
We examine how effective and efficient training with the synthetic datasets from \method and its competitors are.
To evaluate the synthetic datasets, we train MTGNN~\cite{mtgnn} and incorporate our location encoder when required.


Regarding effectiveness, \method consistently outperforms all competitors across compression ratios, as shown in Table \ref{tab:results}, where
we report the mean and standard deviation of the forecasting errors.
Specifically, when the compression ratio is 0.5\%, the model trained on the synthetic data distilled by \method achieves 12\% lower Relative RMSE in the CA dataset compared to the best competitor. 
This demonstrates that \method effectively preserves both temporal and spatial information in a balanced manner within the synthetic dataset.
\vspace{-2mm}

Training the model with the synthetic datasets distilled by \method is also more efficient than training with those distilled by the baselines. As shown in Figure~\ref{fig:comparision},
at a compression ratio of 0.5\%, training on the dataset distilled by \method is up to 6.3$\times$ faster with up to 7.7$\times$ less memory.\footnote{
For clarity, this experiment differs from Preliminary Experiment~\ref{preliminary_experiment} in Section~\ref{intro}, as it employs different models (MTGNN with our location encoder) and training strategies (subset-based training in Algorithm~\ref{algo:train}).
In contrast, Preliminary Experiment~\ref{preliminary_experiment} follows those in \cite{mtgnn}.} These results are consistent with Theorems~\ref{thm:train:time} and~\ref{thm:train:space} in Section~\ref{sec:method:complexity}.

In addition, we evaluate variants of \method that integrate our location encoder and spatial reduction into four baseline distillation methods, denoted as STemDist-[baseline name] in Table~\ref{tab:results}.
These variants outperform their corresponding baselines, showing the effectiveness of our ideas; however, the variants still underperform \method on most datasets.

We further examine when the input is limited to a single feature, spec., the target feature for each dataset. For the GBA, GLA, and CA datasets, this corresponds to their original setup, as they contain only one feature.
As shown in Table~\ref{tab:results}, even under this constraint, \method outperforms most baselines.
\vspace{-2mm}

\subsection{Cross-model Performance}  \label{sec:exp:cross}
We evaluate the cross-model generalization ability of the synthetic datasets, by training widely-used spatio-temporal time series forecasting models: Graph wavenet~\cite{wu2019graph}, STGCN~\cite{stgcn}, and FourierGNN~\cite{fouriergnn}, which are equipped with our location encoder when necessary.\footnote{
We modify STGCN to learn a graph with Eq. (\ref{eq:adj}) without relying on a predefined adjacency matrix. Note that some models (e.g., \cite{fouriergnn,stnorm}) do not require a location encoder when trained on synthetic datasets.}
K-Center~\cite{farahani2009facility}, DC~\cite{zhao2021dataset}, and CondTSF~\cite{ding2024condtsf}, each of which is the best in its respective baseline category, are used as baselines; and the compression ratio is set to 0.5\%.
As shown in Table \ref{tab:cross-archi}, \method leads to the best performance across all tested models and datasets, demonstrating strong cross-model generalizability.
This indicates that the synthetic datasets from \method capture key information that generalizes beyond the surrogate model.
Results for additional models (spec., AGCRN~\cite{bai2020adaptive} and STNORM\cite{stnorm}) are in Appendix E~\cite{appendix}.

\begin{figure}[t]
    \vspace{-1mm}
    \centering
        \includegraphics[width=0.322\linewidth]{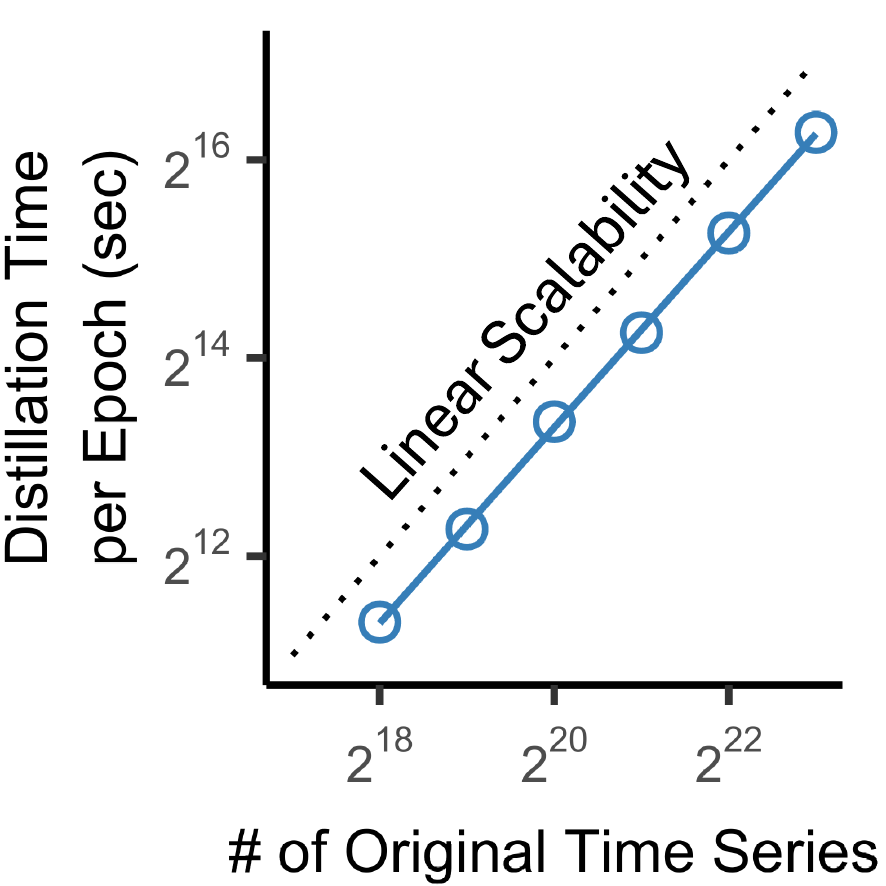}%
    \hspace{-1mm}
        \includegraphics[width=0.322\linewidth]{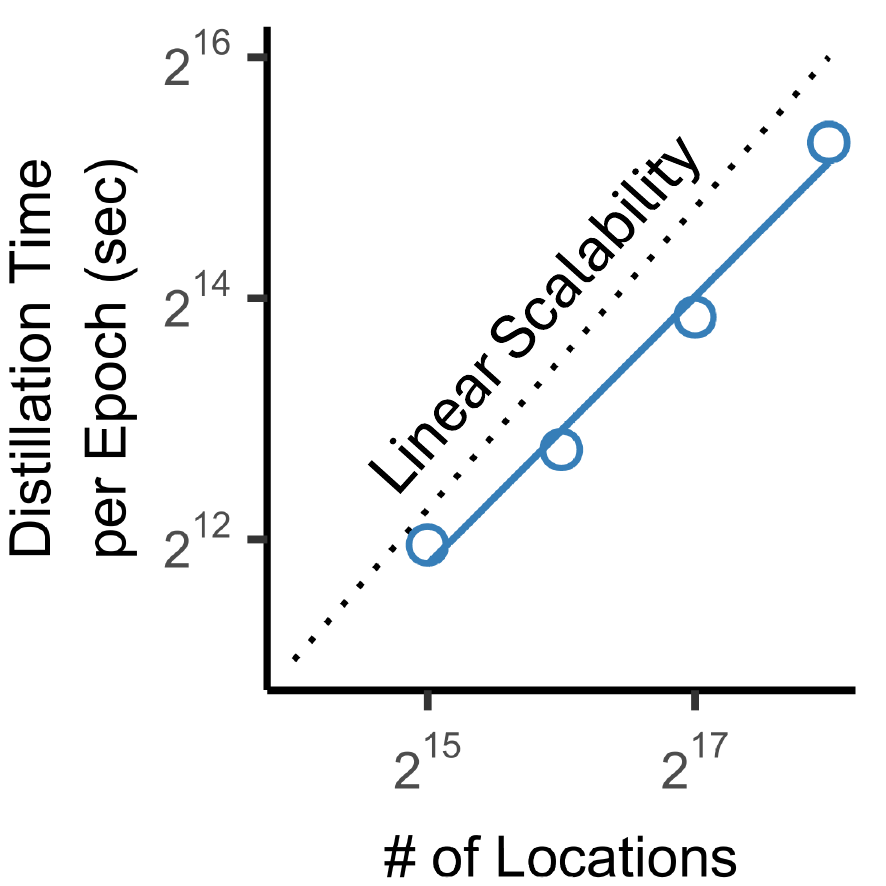}%
    \hspace{-1mm}
        \includegraphics[width=0.322\linewidth]{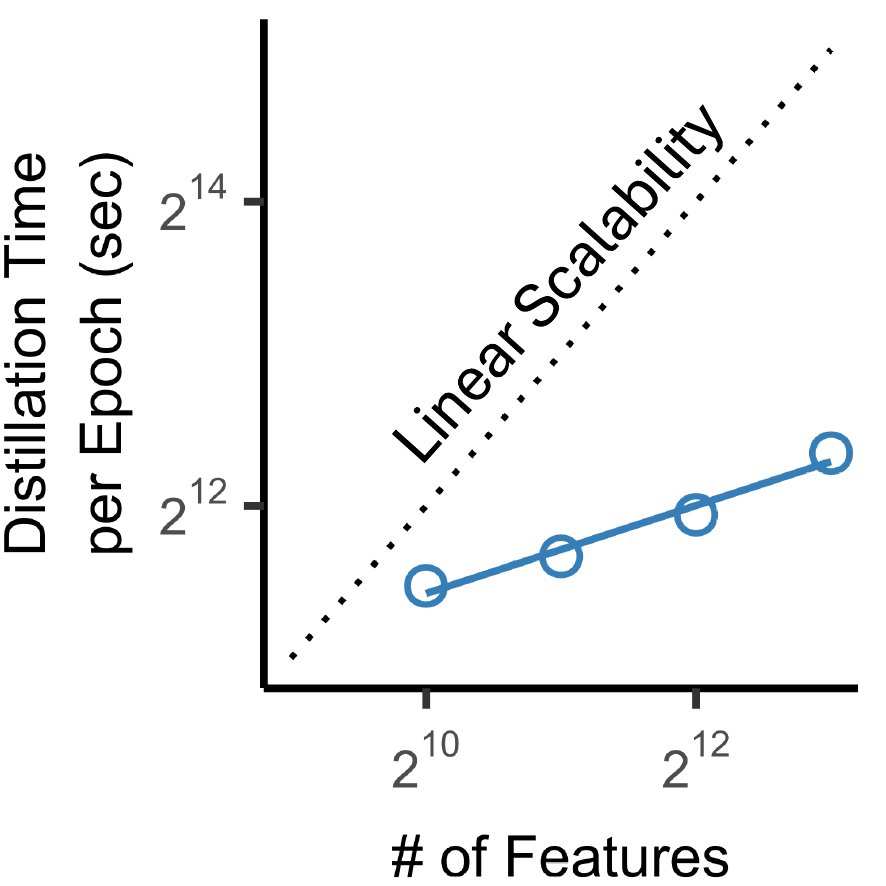}%
   \vspace{-1mm}
    \caption{Scalability of \method. Its empirical distillation time grows (sub-)linearly with (a) the number of original time series, (b) the location count in the original dataset, and (c) the feature count in the original dataset. }
    \label{fig:scalability}
\end{figure}

\subsection{Scalability and Speed}\label{exp:scalability}
\label{sec:exp:distillation_time}

We evaluate the scalability of \method w.r.t. various input/output factors, using artificially generated large-scale datasets.
We progressively double each factor while keeping the others fixed to their default values\footnote{$262,144$ original time series, $64$ synthetic time series, $2,352$ locations, and $1$ feature.}, as summarized in
Table~\ref{tab:dataset}, and measure the per-epoch distillation time required to make a pass over the dataset.
While we control the number of locations in the original dataset, the synthetic dataset size is set accordingly as $N_{\seriessyn} = 0.05 \times N_{\series}$.
Each experiment is conducted three times, and the mean is reported.

As shown in Figure~\ref{fig:scalability}, the distillation time of \method scales linearly (slope of 1 on log–log scale) with the numbers of original time series and locations, and sublinearly (slope less than 1 on log–log scale) with the numbers of features and the number of synthetic time series (see Appendix~F~\cite{appendix} for results with respect to the number of synthetic time series).
The results w.r.t. the number of original time series are consistent with Theorem~\ref{thm:distill} in Section~\ref{sec:method:complexity}.
For other factors, \method shows better empirical scalability than the theoretical upper bounds in Theorem~\ref{thm:distill}, due to the benefits of GPU parallelization within each batch.
Note that parallelization does not take effect when the number of original time series grows, as it increases the number of batches rather than batch size.

\begin{figure}[t] 
    \vspace{-3mm}
    \centering
    \includegraphics[width=0.95\linewidth]{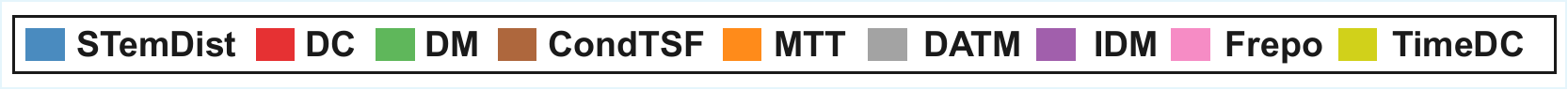} \\
    \centering
    \includegraphics[width=0.90\linewidth]{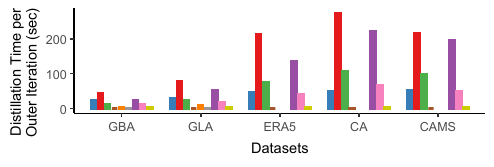} \\
    \vspace{-4mm}
    \caption{Comparison of distillation time per outer iteration across different dataset distillation methods.} \label{fig:distillationtime}
\end{figure}

We also compare the distillation time of \method with the baselines, at a compression ratio of 0.5\%, in Figure \ref{fig:distillationtime}.
The trajectory matching-based approaches (MTT~\cite{cazenavette2022dataset}, DATM~\cite{guo2023towards}, CondTSF~\cite{ding2024condtsf}, and TimeDC~\cite{miao2024less}) generally require less time for distillation.
However, they require additional time to generate expert trajectories prior to the distillation process.
Compared to DM~\cite{zhao2023dataset} and IDM~\cite{zhao2023improved}, \method shows clear time efficiency on datasets with a larger number of locations.
Compared to DC~\cite{zhao2021dataset}, \method consistently shows superior time efficiency across all datasets.

The location encoder incurs negligible inference latency, at most 0.0002 seconds for the real-world datasets (Table~\ref{tab:dataset}) and 0.002 seconds for the artificially generated datasets, accounting for only 0.06\% and 0.004\% of the total inference time.


\subsection{Ablation Study} \label{sec:exp:ablation}
To confirm the effectiveness of each component of \method, we compare \method with three variants:
\begin{itemize}[leftmargin=*]
    \item \textbf{\method w/o G}: \method without subset-based \textbf{G}ranular distillation.
    \item \textbf{\method w/o G \& C}: \method without subset-based \textbf{G}ranular distillation and \textbf{C}lustering of locations.
    \item \textbf{DC}~\cite{zhao2021dataset}: A naive gradient matching method.
\end{itemize}
At a compression ratio of 0.5\%, we perform the distillation three times for each method and evaluate each synthetic dataset three times. We report the mean.



\begin{figure}[t]
\vspace{-1mm}
    \centering
    \includegraphics[width=0.7\linewidth]{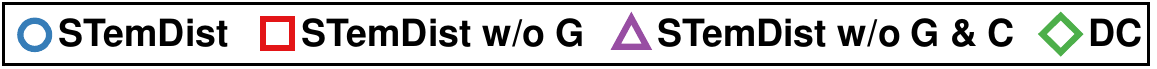} \\
    \vspace{-3mm}
    \subfloat[GLA\label{fig:ablation-gla}]{%
        \includegraphics[width=0.32\linewidth]{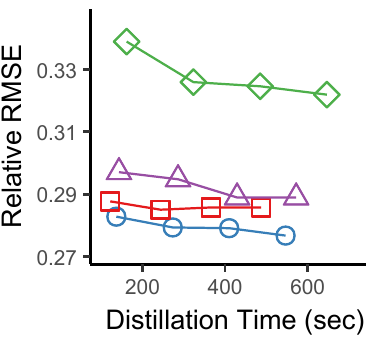}%
    }
    \subfloat[ERA5\label{fig:ablation-era5}]{%
        \includegraphics[width=0.32\linewidth]{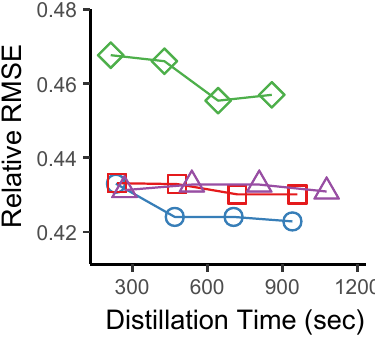}%
    }
    \subfloat[CAMS\label{fig:ablation-cams}]{%
        \includegraphics[width=0.32\linewidth]{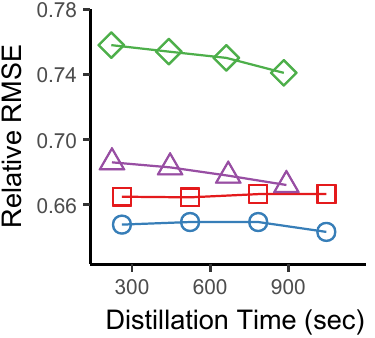}%
    }  
    \caption{Ablation study. Distillation performance of \method and its variants over distillation time on GLA, ERA5, CAMS datasets} \label{fig:ablation}
\end{figure}

Figure~\ref{fig:ablation} shows how the performance of each variant changes over distillation time for GLA, ERA5, and CAMS datasets (refer to Appendix H~\cite{appendix} for results on the other datasets).
Notably, each component of \method helps achieve effective distillation (i.e., lower forecasting error) more rapidly.

We evaluate variants of \method that replace K-means with spatially constrained clustering (RedCap~\cite{guo2008regionalization}) or graph coarsening (METIS~\cite{metis}) for location clustering. For ERA5 and CAMS, we construct grid-based spatial graphs from longitude and latitude coordinates, while traffic datasets are excluded due to missing spatial information. Compared to using K-means (with Relative RMSE of 0.426 on ERA5, 0.653 on CAMS), using METIS (0.421 on ERA5, 0.656 on CAMS) or RedCap (0.429 on ERA5, 0.672 on CAMS) results in comparable distillation performance without clear improvements.

\subsection{Hyperparameter Analysis}
\label{sec:exp:hyperparam}
We examine the effect of hyperparameters on the distillation performance of \method under a fixed total compression ratio of 1\% on the GBA, GLA, and ERA5 datasets. 
Except for the controlled hyperparameters, all others are fixed to the default values described in Section~\ref{sec:exp:settings}.
We evaluate four temporal–spatial ratio combinations and observe that \method tends to be more effective when the temporal and spatial ratios are balanced (Table~\ref{tab:sensitivity}(a)). We further examine the impact of the number of subsets $K$ in subset-based granular distillation. 
As shown in Table~\ref{tab:sensitivity}(b), \method performs similarly well when the number of subsets is four or greater.
Recall that, in Section~\ref{sec:exp:ablation}, we observe a substantial performance drop when the number of subsets is one (\method w/o G in Figure~\ref{fig:ablation}).
Hyperparameter analysis results under a total compression ratio of 0.5\% are provided in Appendix~I~\cite{appendix}.

\begin{table}[t]
\vspace{-3mm}
\centering
\caption{Effect of hyperparameters on the distillation performance of \method, measured by Relative RMSE under a total compression ratio of 1\%.} \label{tab:sensitivity}
\par \vspace{-3mm}
\subfloat[Effect of temporal and spatial compression ratios]{
\resizebox{\linewidth}{!}{
\begin{tabular}{l || c c c c}
\toprule
& \multicolumn{4}{c}{
Temporal--Spatial Compression Pairs $(\frac{M_\seriessyn}{M_\series}, \frac{N_\seriessyn}{N_\series})$} \\
\cmidrule(lr){2-5}
Dataset 
& (2\%, 50\%) & (5\%, 20\%) & (10\%, 10\%) & (25\%, 4\%) \\
\midrule
GBA & 0.319 $\pm$ 0.008 & 0.307 $\pm$ 0.004 & 0.307 $\pm$ 0.002 & 0.324 $\pm$ 0.007 \\
GLA & 0.291 $\pm$ 0.004 & 0.278 $\pm$ 0.002 & 0.272 $\pm$ 0.001 & 0.279 $\pm$ 0.003\\
ERA5 & 0.426 $\pm$ 0.004 & 0.424 $\pm$ 0.003 & 0.426 $\pm$ 0.004 & 0.424 $\pm$ 0.003\\
CA  & 0.302 $\pm$ 0.009 & 0.292 $\pm$ 0.002 & 0.287 $\pm$ 0.002 & 0.293 $\pm$ 0.003 \\
CAMS  & 0.656 $\pm$ 0.007 & 0.652 $\pm$ 0.003 & 0.652 $\pm$ 0.006 & 0.653 $\pm$ 0.003 \\
\midrule
Avg. & 0.399 $\pm$ 0.006 & 0.391 $\pm$ 0.003 & 0.389 $\pm$ 0.003 & 0.395 $\pm$ 0.004 \\
\bottomrule
\end{tabular}
}}

\par\vspace{-2mm}

\subfloat[Effect of the number of subsets]{
 \resizebox{\linewidth}{!}{ 
\begin{tabular}{l || c c c c} \toprule & \multicolumn{4}{c}{Number of Subsets $K$} 
\\ \cmidrule(lr){2-5} Dataset & 2 & 4 & 8 & 16 \\ 
\midrule 
GBA & 0.352 $\pm$ 0.007 & 0.307 $\pm$ 0.002 & 0.313 $\pm$ 0.009 & 0.310 $\pm$ 0.011 \\ 
GLA & 0.274 $\pm$ 0.006 & 0.272 $\pm$ 0.001 & 0.275 $\pm$ 0.004 & 0.276 $\pm$ 0.004 \\ 
ERA5 & 0.425 $\pm$ 0.005 & 0.426 $\pm$ 0.004 & 0.434 $\pm$ 0.005 & 0.429 $\pm$ 0.003 \\ 
CA & 0.297 $\pm$ 0.007 & 0.287 $\pm$ 0.002 & 0.281 $\pm$ 0.002 & 0.280 $\pm$ 0.006 \\ 
CAMS & 0.664 $\pm$ 0.003 & 0.652 $\pm$ 0.006 & 0.652 $\pm$ 0.003 & 0.654 $\pm$ 0.003 \\  
\midrule
Avg. & 0.402 $\pm$ 0.006 & 0.389 $\pm$ 0.003 & 0.391 $\pm$ 0.006 & 0.390 $\pm$ 0.005\\
\bottomrule 
\end{tabular}  
}}
\end{table}

\section{Conclusions and Future Directions}\label{conclusion}
In this work, we propose \method, a novel dataset distillation method specialized for spatio-temporal time series.
The main idea of \method is bi-dimensional compression, which simultaneously reduces both the temporal and spatial dimensions in the synthetic dataset, enabling more effective and efficient model training. 
This is done by three core components (location encoders, location clustering, and subset-based granular distillation), which enhance both effectiveness and efficiency of the distillation process. 
In our experiments, the synthetic data generated by \method makes model training \textbf{faster} (up to 6$\times$), more \textbf{memory-efficient} (up to 8$\times$), and more \textbf{effective} (with up to 12\% lower prediction error). 
A potential future direction is to extend \method in a cost-sensitive manner to prioritize performance preservation for high-stakes or rare events (e.g., extreme weather events in meteorological data).

\smallsection{Acknowledgements}
This work was supported by Institute of Information \& communications Technology Planning \& Evaluation (IITP) grant funded by the Korea government (MSIT) (No. RS-2024-00457882, AI Research Hub Project, 40\%) (No. RS-2022-II220157, Robust, Fair, Extensible Data-Centric Continual Learning, 40\%) (No. RS-2024-00438638, EntireDB2AI: Foundations and Software for Comprehensive Deep Representation Learning and Prediction on Entire Relational Databases, 10\%)
(RS-2019-II190075, Artificial Intelligence Graduate School Program (KAIST), 10\%).

\newpage

\section*{AI-Generated Content Acknowledgement}

Portions of the manuscript were revised with assistance from ChatGPT (GPT-5, OpenAI) for language refinement. The authors reviewed and edited all AI-generated content and take full responsibility for the final version of the manuscript.

\bibliographystyle{IEEEtran}
\bibliography{sample}

\end{document}